%% file: main.tex
\documentclass{article}

\usepackage{microtype}
\usepackage{graphicx}
\usepackage{subfigure}
\usepackage{booktabs} % for professional tables

\usepackage{hyperref}

\usepackage[accepted]{icml2023}

\usepackage{amsmath}
\usepackage{amssymb}
\usepackage{mathtools}
\usepackage{amsthm}

\usepackage[capitalize,noabbrev]{cleveref}

\theoremstyle{plain}

\theoremstyle{definition}

\theoremstyle{remark}

\usepackage{hyperref}
\usepackage{url}
\usepackage{amsmath}
\usepackage{bm}
\usepackage{bbm}
\usepackage{xspace}
\usepackage{graphicx}
\usepackage{comment}
\usepackage{multirow}

\usepackage{rotating}

\usepackage{amssymb}

\usepackage{xcolor}

\newcommand*{\ie}{i.e.\@\xspace}
\newcommand*{\eg}{e.g.\@\xspace}
\newcommand*{\etc}{etc.\@\xspace}

\newcommand{\indic}[1]{\mathbbm{1}_{[#1]}}

\usepackage[textsize=tiny]{todonotes}

\icmltitlerunning{Spatial Implicit Neural Representations for Global-Scale Species Mapping}

\begin{document}

\twocolumn[
\icmltitle{Spatial Implicit Neural Representations for Global-Scale Species Mapping}

\icmlsetsymbol{equal}{*}

\begin{icmlauthorlist}
\icmlauthor{Elijah Cole}{yyy}
\icmlauthor{Grant Van Horn}{zzz}
\icmlauthor{Christian Lange}{eee}
\icmlauthor{Alexander Shepard}{nnn}
\icmlauthor{Patrick Leary}{nnn}
\icmlauthor{Pietro Perona}{yyy}
\icmlauthor{Scott Loarie}{nnn}
\icmlauthor{Oisin Mac Aodha}{eee}
\end{icmlauthorlist}

\icmlaffiliation{yyy}{Caltech}
\icmlaffiliation{zzz}{Cornell}
\icmlaffiliation{nnn}{iNaturalist}
\icmlaffiliation{eee}{University of Edinburgh}

\icmlcorrespondingauthor{Elijah Cole}{ecole@caltech.edu}

\icmlkeywords{species distribution modeling, coordinate networks, neural networks}

\vskip 0.3in
]



\printAffiliationsAndNotice{}  % leave blank if no need to mention equal contribution

\begin{abstract}
Estimating the geographical range of a species from sparse observations is a challenging and important geospatial prediction problem. Given a set of locations where a species has been observed, the goal is to build a model to predict whether the species is present or absent at any location. This problem has a long history in ecology, but traditional methods struggle to take advantage of emerging large-scale crowdsourced datasets which can include tens of millions of records for hundreds of thousands of species. In this work, we use Spatial Implicit Neural Representations (SINRs) to jointly estimate the geographical range of 47k species simultaneously. We find that our approach scales gracefully, making increasingly better predictions as we increase the number of species and the amount of data per species when training. To make this problem accessible to machine learning researchers, we provide four new benchmarks that measure different aspects of species range estimation and spatial representation learning. Using these benchmarks, we demonstrate that noisy and biased crowdsourced data can be combined with implicit neural representations to approximate expert-developed range maps for many species.
\end{abstract}
 
\section{Introduction}

We are currently observing a dramatic decline in global biodiversity, which has severe ramifications for natural resource management, food security, and ecosystem services that are crucial to human health~\citep{watson2019summary,rosenberg2019decline}. In order to take effective conservation action we must understand species' ranges, \ie where they live. However, we only have estimated ranges for a relatively small number of species in limited areas, many of which are already out of date by the time they are released. 

The range of a species is typically estimated through \emph{Species Distribution Modeling} (SDM) \citep{elith2009species}, the process of using species observation records to develop a statistical model for predicting whether a species is present or absent at any location. 
With enough \emph{presence-absence} data (\ie records of where a species has been confirmed to be present and absent) this problem can be approached using standard statistical learning methods~\citep{beery2021species}.\footnote{The term ``presence-absence" should not be taken to convey absolute certainty about whether a species is present or absent. False absences (\ie non-detections) and, to a lesser extent, false presences are a serious concern in SDM \citep{mackenzie2002estimating}.} 
However, presence-absence data is scarce due to the difficulty of verifying that a species is truly absent from an area. \emph{Presence-only} data (\ie verified observation locations, with no confirmed absences) is much more abundant as it is easier to collect. For instance, the community science platform iNaturalist~\citep{iNatWeb} has collected over 141M presence-only observations to date across 429k species. Though presence-only data is not without drawbacks~\citep{hastie2013inference}, it is important to develop methods that can take advantage of this vast supply of data.

Deep learning is one of our best tools for making use of large-scale datasets. Deep neural networks also have a key advantage over many existing SDM methods because they can \emph{jointly} learn the distribution of many species in the same model~\citep{chen2017deep,tang2018multi,mac2019presence}. By learning representations that share information across species, the models can make improved predictions~\citep{chen2017deep}. However, the majority of current deep learning approaches need presence-absence data for training, which prevents them from scaling beyond the small number of species and regions for which sufficient presence-absence data is available. 

\begin{figure*}[t]
\centering
\includegraphics[width=1.0\textwidth]{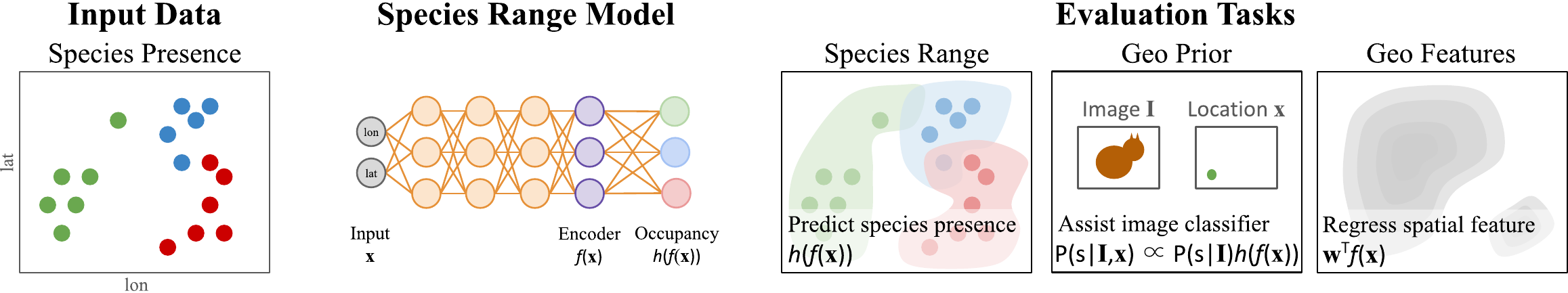}
\vspace{-20pt}
\caption{
    We show that sparse species observation data can be used to train Spatial Implicit Neural Representations (SINRs) which are transferable to other geospatial tasks. \emph{(Left)} Here we show sparse, presence-only, spatial observations for three toy species (red, green, and blue). \emph{(Middle)} The species observations are used to train a neural network that consists of a spatial feature encoder and per-species presence predictors. \emph{(Right)} We evaluate on three diverse tasks: (i) estimating species ranges, (ii) assisting image  classifiers using geographical range priors, and (iii) regressing geospatial features via our learned SINR.
}
\vspace{-8pt}
\label{fig:overview}
\end{figure*}

Our work makes the following contributions: \\
(i) We show that implicit neural representations trained with noisy crowdsourced presence-only data can be used to estimate dense species' ranges.  We call these models Spatial Implicit Neural Representations (SINRs).\footnote{We slightly abuse the terminology by using ``SINR" to refer to both the model and the representation it parameterizes.}\\
(ii) We conduct a detailed investigation of loss functions for learning from presence-only data, their scaling properties, and the resulting geospatial representations. \\
(iii) We provide a suite of four geospatial benchmark tasks -- ranging from species mapping to fine-grained image classification -- which will facilitate future research on spatially sparse high-dimensional implicit neural representations, large-scale SDM, and geospatial representation learning. 

Training and evaluation code is available at:
\begin{center}\vspace{-4mm}
\url{https://github.com/elijahcole/sinr}
\end{center}

\section{Related Work}

Species distribution modeling (SDM) refers to a set of methods that aim to predict where (and sometimes when, and in what quantities) species of interest are likely to be found~\citep{elith2009species}. The literature on SDM is vast. Readers interested in an overview  should consult the review by~\citet{elith2009species} or the recent review of SDM for computer scientists by~\citet{beery2021species}. Note that we focus narrowly on the problem of predicting the occurrence of a species at a location, \ie we do not consider more complex problems like trend or abundance estimation~\citep{potts2006comparing}.

Traditional approaches to SDM train conventional supervised learning models (\eg logistic regressors~\citep{pearce2000evaluation}, random forests~\citep{cutler2007random}, \etc) to learn a mapping between hand-selected sets of environmental features (\eg altitude, average rainfall, \etc) and species presence or absence~\citep{phillips2004maximum,elith2006novel}. Readers interested in these approaches should consult~\citet{norberg2019comprehensive, valavi2021modelling, valavi2022predictive}, and the references therein. More recently, deep learning methods have been introduced that instead \emph{jointly} represent multiple different species within the same model~\citep{chen2017deep,botella2018deep,tang2018multi,mac2019presence,teng2023bird}. These models are typically trained on crowdsourced data, which can introduce additional challenges and biases that need to be accounted for during training~\citep{fink2010spatiotemporal,chen2019bias,johnston2020estimating,botella2021jointly}. We build on the work of \citet{mac2019presence}, who proposed a neural network approach that forgoes the need for environmental features (as used by \eg~\citet{botella2018deep,tang2018multi}) by learning to predict species presence from geographical location alone. 

The problem of joint SDM with presence-only data can be viewed as an instance of multi-label classification with incomplete supervision. In particular, it is an example of Single Positive Multi-Label (SPML) learning ~\citep{cole2021multi,verelst2022spatial,zhou2022acknowledging}. The goal is to train a model that is capable of making multi-label predictions at test time, despite having only ever observed one positive label per training instance (\ie no confirmed negative training labels). Our work connects the SPML literature and SDM literature, and sets up large-scale joint species distribution modeling as a challenging real-world SPML task. This setting presents significant new difficulties for SPML, which has largely been limited to artificial label bias patterns~\citep{arroyo2023understanding} and relatively small label spaces ($<100$ categories). 
Some SPML methods such as ROLE~\citep{cole2021multi} are not computationally viable when the label space is large. One of our baselines is based on the SPML method of~\citet{zhou2022acknowledging}, which is scalable and obtains nearly state-of-the-art performance on the standard SPML benchmarks~\citep{cole2021multi}, but it is not a top performer on our new benchmark tasks. 

Our work is related to the growing number of papers that use coordinate neural networks for implicitly representing images~\citep{tancik2020fourier} and 3D scenes~\citep{sitzmann2019scene,mildenhall2020nerf}. There are many design choices in these methods that are being actively studied, including the impact of the activation functions in the network~\citep{sitzmann2019scene,ramasinghe2021beyond} and the effect of different input encodings~\citep{tancik2020fourier,zheng2022trading}. In most research on implicit neural representations, there is an obvious choice of training objective, \eg mean squared error between the predictions and the data. In the context of presence-only species estimation, this choice is less clear. We systematically investigate this question in our experiments. Our benchmark also facilitates investigations of implicit neural representations with high-dimensional output spaces and sparse supervision. 

Quantifying the performance of SDM at scale is notoriously difficult due to the fact that we lack confirmed presence-absence data for most species and locations~\citep{beery2021species}. 
One approach is to evaluate performance on a small set of species from limited geographical regions where it is feasible to collect presence-absence data, as done in \eg~\citet{potts2006comparing, norberg2019comprehensive, valavi2022predictive}. Two of our evaluation tasks are larger-scale versions of this idea, in which we compare the performance of our models against expert range maps. An alternative evaluation approach is to measure the performance on a related ``proxy'' task. For example, there have been a number of works that use models trained for species range estimation to assist deep image classifiers~\citep{berg2014birdsnap,tang2015improving,mac2019presence,chu2019geo,mai2019multi,terry2020thinking,skreta2020spatiotemporal,yang2022dynamic}. By using images from platforms like iNaturalist, we can evaluate different range estimation methods on the task of aiding fine-grained image classification across tens of thousands of species.  Finally, we also evaluate the spatial representations learned by our models via transfer learning, using them as inputs for a set of geospatial regression tasks. These complementary benchmark tasks capture different aspects of performance, and provide a starting point for large-scale SDM evaluation. 
See  Figure~\ref{fig:overview} for an overview of our tasks. 

\section{Methods}

\subsection{Preliminaries}\label{sec:prelim}

\noindent
\textbf{Problem statement.} Let $\mathbf{x} = [lon, lat]$ denote a geographical location (\ie longitude and latitude). Let $\mathbf{y} \in \{0, 1\}^S$ denote the true presence ($1$) or absence ($0$) of $S$ different species at location $\mathbf{x}$. Following~\citet{cole2021multi}, we introduce $\mathbf{z} \in \{0, 1, \varnothing\}^S$ to represent our observed data at $\mathbf{x}$, where $z_{j} = 1$ if species $j$ is present, $z_{j} = 0$ if species $j$ is absent, and $z_{j} = \varnothing$ if we do not know whether species $j$ is present or absent. Our goal is to develop a model that produces an estimate of $\mathbf{y}$ at any location $\mathbf{x}$ over some spatial domain $\mathcal{X}$, given observed data $\{(\mathbf{x}_i, \mathbf{z}_i)\}_{i=1}^N$. We parameterize this model as $\hat{\bm{y}} = h_\phi(f_\theta(\bm{x}))$, where $f_\theta: \mathcal{X} \to \mathbb{R}^k$ is a location encoder with parameters $\theta$ and $h_\phi: \mathbb{R}^k \to [0, 1]^S$ is a multi-label classifier with parameters $\phi$. 
The prediction $\hat{\bm{y}} \in [0, 1]^S$ is our estimate of how likely each species is to be present at $\mathbf{x}$. 

\begin{figure*}[t]
\centering
\includegraphics[trim={0pt 100pt 0pt 0pt},clip, width=0.49\textwidth]{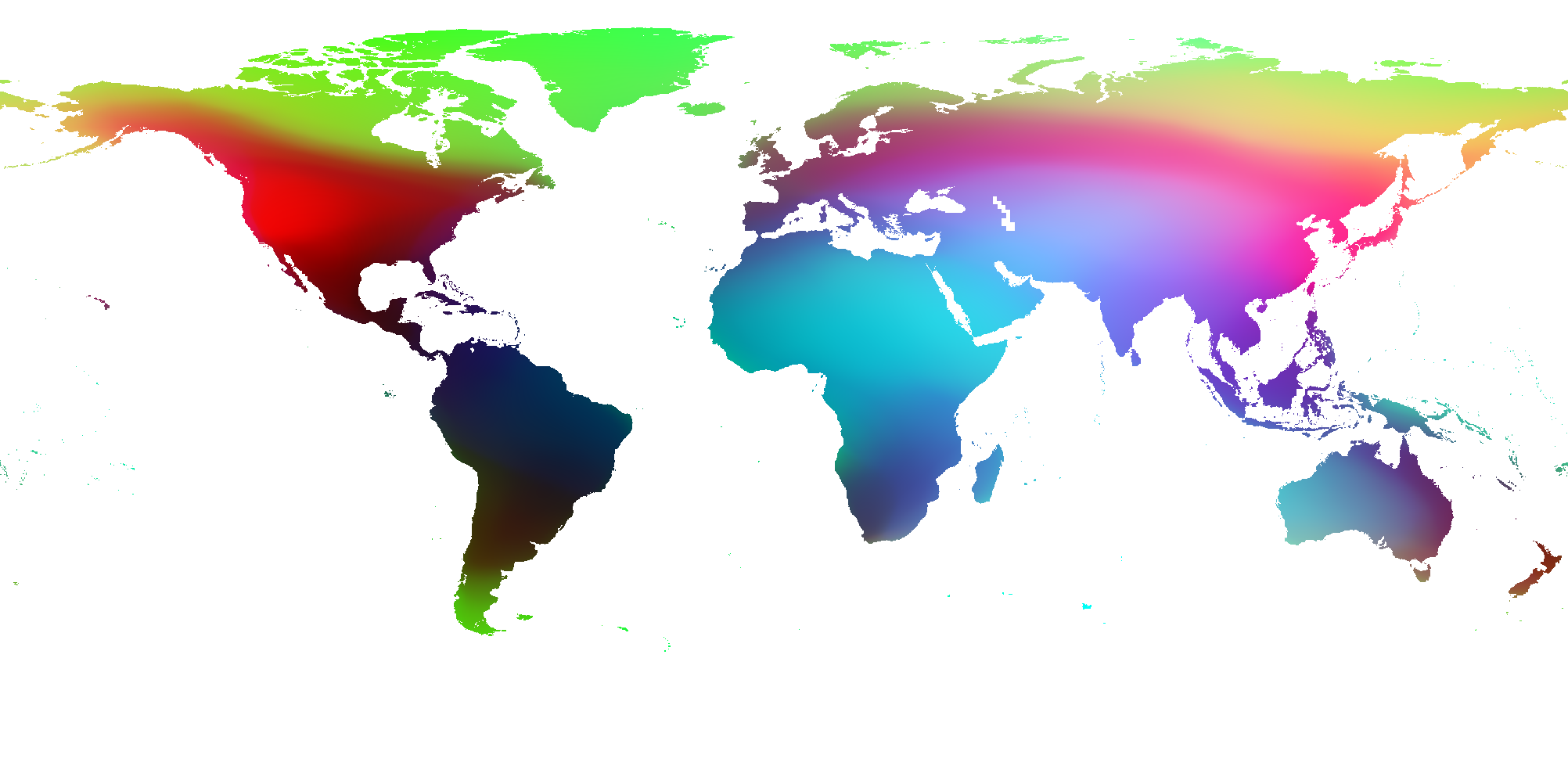} \hfill
\includegraphics[trim={0pt 100pt 0pt 0pt},clip, width=0.49\textwidth]{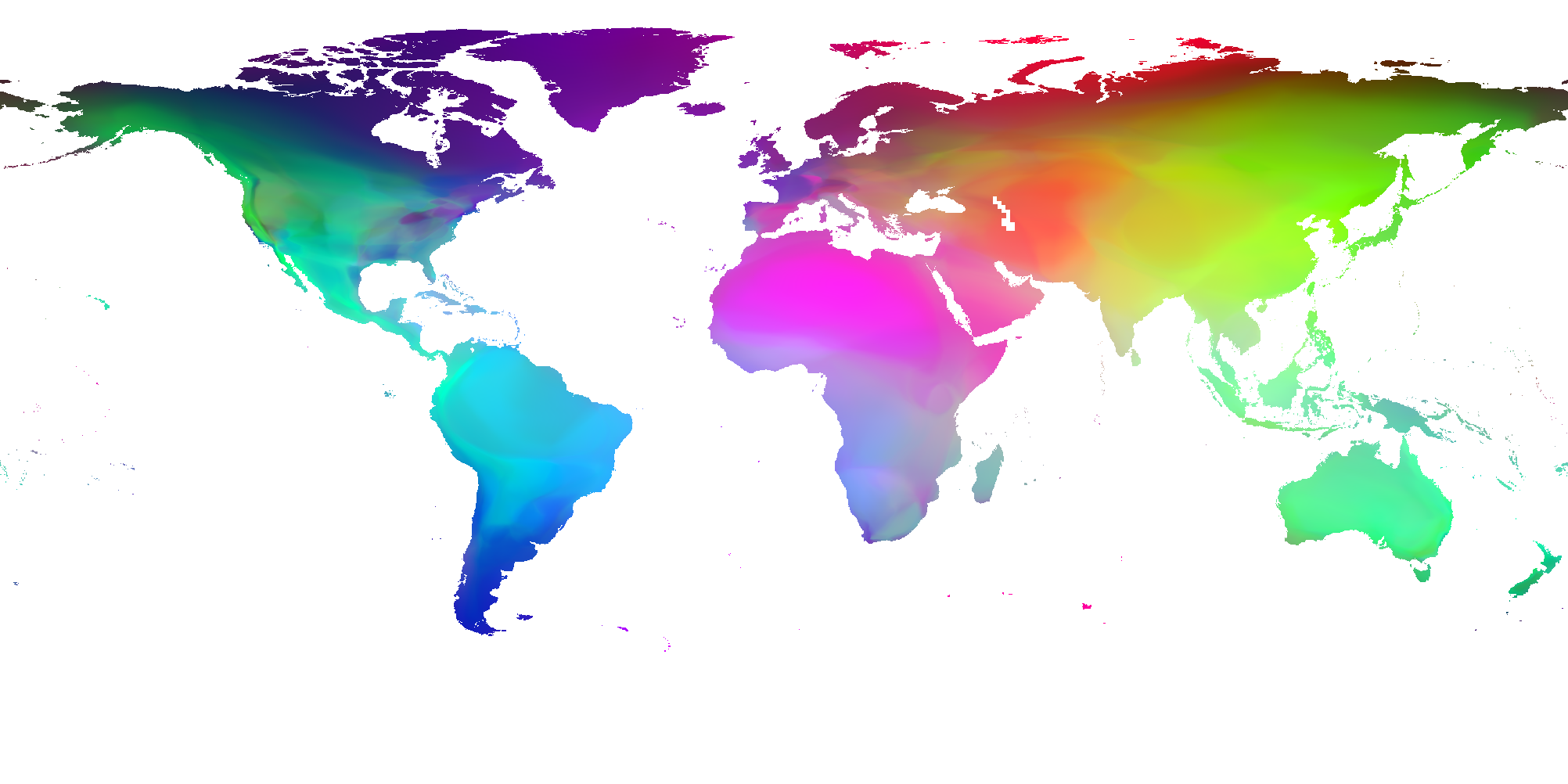}
\vspace{-10pt}
\caption{
    Visualization of the $256$-dimensional features from learned location encoders $f_\theta$ projected to three dimensions using Independent Component Analysis (ICA). All models use the $\mathcal{L}_\mathrm{AN-full}$ loss and take coordinates as input. \emph{(Left)} This corresponds to a SINR model trained with a maximum of {\it 10} examples per class. The features are smooth and do not appear to encode much high frequency spatial information. \emph{(Right)} In contrast, the SINR model trained with a maximum of {\it 1000} examples per class contains more high frequency information. The increase in training data appears to enable this model to better encode spatially varying environmental properties. Note, ICA is performed independently per-model, so similar colors do not indicate correspondence between the two images. 
}
\label{fig:geo_ica}
\end{figure*}

Intuitively, the location encoder $f_\theta$ provides a representation of geographical space that is used by the multi-label classifier $h_\phi$ to predict species presence at each location. If $\theta$ is fixed or if $f$ is a differentiable function of $\theta$, then we can use standard methods like stochastic gradient descent to approximately solve 
\begin{align}
    \theta^*, \phi^* = \mathrm{argmin}_{\theta, \phi} \frac{1}{N}\sum_{i=1}^N \mathcal{L}(\hat{\mathbf{y}_i}, \mathbf{z}_i)
\end{align}
where $\hat{\bm{y}}_i = h_\phi(f_\theta(\bm{x}_i))$ and $\mathcal{L}$ is a suitably chosen loss function.
Once trained, we say that $h_\phi \circ f_\theta$ has learned a Spatial Implicit Neural Representation (SINR) for the distribution of each species in the training set. Along the way we can learn $f_\theta$, which produces a representation for any location on earth. See Figure~\ref{fig:geo_ica} for visualizations of some of these geospatial representations. 

\noindent
\textbf{Input encoding.} Each species observation is associated with spatial coordinates $\mathbf{x} = [lon, lat]$. 
In practice, we rescale these values so that $lon, lat \in [-1, 1]$ and, following \citet{mac2019presence}, we guard against boundary effects using a sinusoidal encoding. 
The results is an input vector 
\begin{align}\label{eq:input_enc}
    \mathbf{x} = \left[\sin(\pi~lon), \cos(\pi~lon), \sin(\pi~lat), \cos(\pi~lat)\right].
\end{align} 
Alternative input encodings for related coordinate networks have been explored in the existing literature~ \citep{mai2019multi,tancik2020fourier,mai2022sphere2vec,zheng2022trading}. 
This choice is orthogonal to the losses we explore, so we leave the evaluation of input encodings to future work. 

\noindent
\textbf{Implicit neural representations.} Traditionally, representation learning aims to transform complex objects (\eg images, text) into simpler objects (\eg low-dimensional vectors) that facilitate downstream tasks like classification or regression~\citep{Goodfellow-et-al-2016}. Implicit neural representations offer a different perspective, in which a signal is represented by a neural network that maps the signal domain (\eg $\mathbb{R}$ for audio, $\mathbb{R}^2$ for images) to the signal values~\citep{sitzmann2019scene, tancik2020fourier}. In this work we learn implicit neural representations from a large collection of crowdsourced data containing observations of many species. This yields an implicit neural representation for the geospatial distribution of each species, as well as a representation for any location on earth.

\noindent
\textbf{Presence-absence vs. presence-only data.} Species observation datasets come in two varieties: (i) \emph{Presence-absence} data consists of locations where a species has been observed to be present and locations where it has been confirmed to be absent. That is, we say we have presence-absence data for species $j$ if $|\{\mathbf{z}_i: z_{ij} = 0\}| > 0$ and $|\{\mathbf{z}_i: z_{ij} = 1\}| > 0$. 
Unfortunately, presence-absence data is costly to obtain at scale because confirming absence requires skilled observers to exhaustively search an area. (ii) \emph{Presence-only} data is easier to acquire and thus more abundant because absences are not collected, \ie $z_{ij} \in \{1, \varnothing \}$, for $i\in [N]$ and $j\in [S]$. 

\subsection{Learning from Large-Scale Presence-Only Data}
\label{sec:main_losses}

\begin{figure}[t]
\centering
\includegraphics[width=0.48\textwidth]{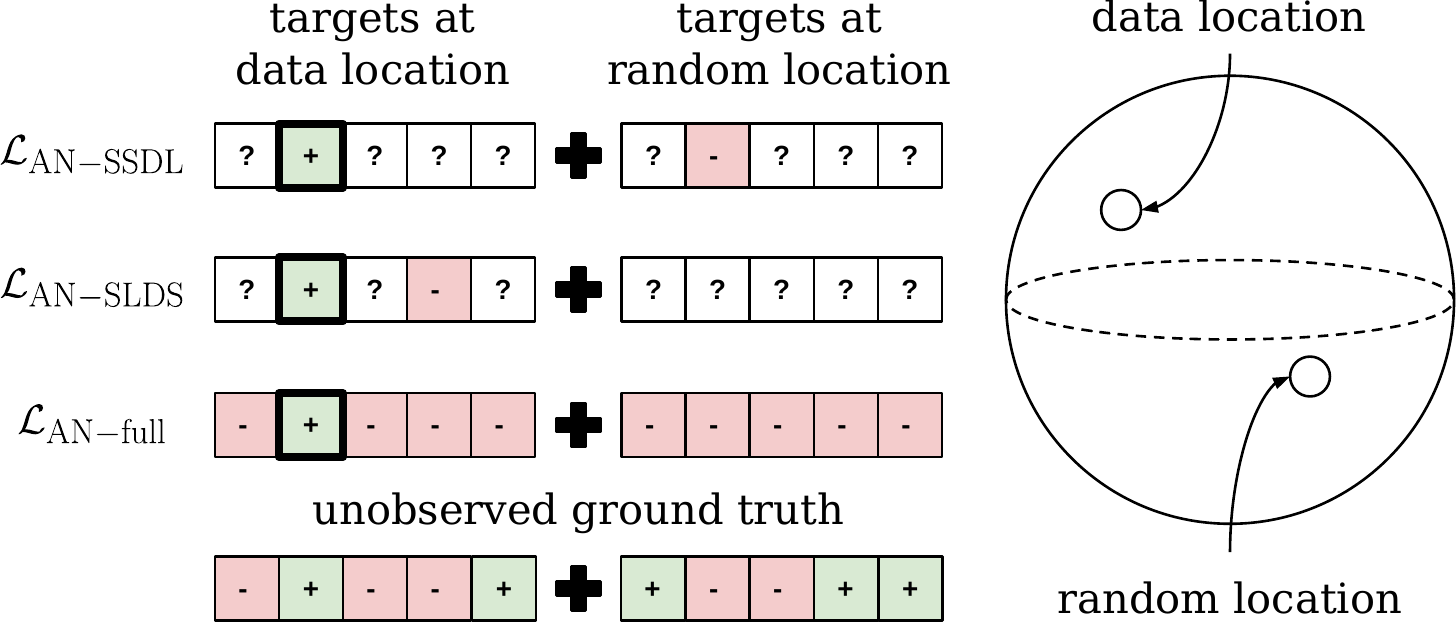}
\vspace{-14pt}
\caption{
    Illustration of the data used by three loss functions from Section~\ref{sec:main_losses}. For each loss,  we visualize the targets that the network is trained to predict. 
    Each loss can be broken into two parts: one part that updates the network's predictions at the location of a training example (\emph{data location}) and one part that updates the network's predictions at another location chosen randomly (\emph{random location}). Each loss has access to one confirmed positive label (bold boxes). The rest of the labels are unobserved (non-bold boxes), and the losses make different, imperfect, assumptions about those unobserved labels. 
}
\label{fig:loss_illustration}
\vspace{-5pt}
\end{figure}

In the context of training SPML \emph{image} classifiers, a simple but effective approach is to assume that unobserved labels are negative \citep{cole2021multi}. This approach is based on a probabilistic argument: since natural images tend to contain a small number of categories compared to the size of the label set, the vast majority of the labels will be negative. This is also true for species distribution modeling. Given an arbitrary location and a large set of candidate species, nearly all of them will be absent. In this section we describe several simple and scalable loss functions based on this idea. We illustrate three of our  losses in Figure~\ref{fig:loss_illustration}.

\textbf{``Assume negative" loss (same species, different location).} As confirmed absences are not available in the presence-only setting, a common approach is to use randomly generated ``pseudo-negatives'' \citep{phillips2009sample}.    
This first loss pairs each observation of a species with a pseudo-negative for that species at another location chosen uniformly at random:
\begin{align}
    \mathcal{L}_\mathrm{AN-SSDL}(\hat{\mathbf{y}}, \mathbf{z}) =  &-\frac{1}{n_\mathrm{pos}}\sum_{j=1}^S \indic{z_{j}=1}[ \log(\hat{y}_{j})\\
    &+ \log(1 - \hat{y}_j')] \nonumber
\end{align}
where $\hat{\mathbf{y}}' = h_\phi(f_\theta(\mathbf{r}))$ with $\mathbf{r} \sim \mathrm{Uniform}(\mathcal{X})$ and $n_\mathrm{pos} = \sum_{j=1}^S \indic{z_j = 1}$. 
This approach generates pseudo-negatives (\ie random absences) across the globe, but many of them are likely to be ``easy'' because they are far from the true species range. 

\textbf{``Assume negative" loss (same location, different species).} This loss pairs each observation of a species with a pseudo-negative at the same location for a different species:
\begin{align} 
\mathcal{L}_\mathrm{AN-SLDS}(\hat{\mathbf{y}}, \mathbf{z}) &=  - \frac{1}{n_\mathrm{pos}}\sum_{j=1}^S \indic{z_{j}=1} [\log(\hat{y}_j)\\
& \quad + \log(1 - \hat{y}_{j'}) \nonumber]
\end{align}
where $j' \sim \mathrm{Uniform}(\{j : z_{j} \neq 1\})$. Intuitively, this approach generates pseudo-negatives that are aligned with the spatial distribution of the observed data. 

\noindent
\textbf{Full ``assume negative" loss.} The previous two losses are inefficient in the sense that they do not use all of the entries in $\hat{\mathbf{y}}$. 
We can combine the pseudo-negative sampling strategies of $\mathcal{L}_\mathrm{AN-SSDL}$ and $\mathcal{L}_\mathrm{AN-SLDS}$ and use all available predictions as follows:
\begin{align}
\mathcal{L}_\mathrm{AN-full}(\hat{\mathbf{y}}, \mathbf{z}) &= -\frac{1}{S} \sum_{j=1}^S [\indic{z_j=1}\lambda \log(\hat{y}_j)\\
&  \quad  + \indic{z_j \neq 1}\log(1-\hat{y}_j) + \left. \log(1-\hat{y}_j') \nonumber \right]
\end{align}
where $\hat{\mathbf{y}}' = h_\phi(f_\theta(\mathbf{r}))$ with $\mathbf{r} \sim \mathrm{Unif}(\mathcal{X})$. The hyperparameter $\lambda > 0$ can be used to prevent the negative labels from dominating the loss.
This is equivalent to the loss from \citet{mac2019presence}, but without their user modeling terms. Their version (including user modeling terms) is $\mathcal{L}_\mathrm{GP}$ in Table~\ref{tab:loss_benchmark} (``GP" = ``Geo Prior"). 

\noindent
\textbf{Maximum entropy loss.}  \citet{zhou2022acknowledging} recently proposed a simple but effective and scalable technique for SPML image classification. Their approach encourages predictions for unobserved labels to maximize entropy instead of forcing them to zero like the ``assume negative" approaches we have been discussing. 
We can apply this idea to $\mathcal{L}_\mathrm{AN-SSDL}$, $\mathcal{L}_\mathrm{AN-SLDS}$, and $\mathcal{L}_\mathrm{AN-full}$ by replacing all terms of the form ``$-\log(1-p)$'' with terms of the form ``$H(p)$'', where $H(p) = - (p \log(p) + (1-p) \log(1-p))$ is the Bernoulli entropy. We write these ``maximum entropy" (ME) variants as $\mathcal{L}_\mathrm{ME-SSDL}$,  $\mathcal{L}_\mathrm{ME-SLDS}$, and $\mathcal{L}_\mathrm{ME-full}$. 
(\citet{zhou2022acknowledging} also includes a pseudo-labeling component, but we omit this because \citet{zhou2022acknowledging} shows that it provides only a small improvement.) 

\section{Experiments}
In this section we investigate the performance of SINR models on four species and environmental prediction tasks. 

\subsection{Models}

As described in Section~\ref{sec:prelim}, our SINR models consist of a location encoder $f_\theta$ and a multi-label classifier $h_\phi$ which produce a vector of predictions $\hat{y} = h_\phi(f_\theta(\mathbf{x}))$ for a location $\mathbf{x}$. The location encoder $f_\theta$ is implemented as the fully connected neural network shown in Figure~\ref{fig:arch}. We implement the multi-label classifier $h_\phi$ as a single fully connected layer with sigmoid activations. 
For fair companions, we follow a similar architecture to \citet{mac2019presence}. 
Full implementation details can be found in Appendix~\ref{sec:impl_det}. 

Besides SINR, we study two other model types. The first is logistic regression \citep{pearce2000evaluation}, in which the location encoder $f_\theta$ is replaced with the identity function and $h_\phi$ is unchanged. Logistic regression is commonly used for SDM in the ecology literature. It also has the virtue of being highly scalable since it can be trained using GPU-accelerated batch-based optimization. The second type of non-SINR model is the discretized grid model. These models do not use a location encoder at all, but instead make predictions based on binning the training data \citep{berg2014birdsnap}. Full details for these models can be found in Appendix~\ref{sec:impl_det}. These baselines allow us to quantify the importance of the deep location encoder in our SINR models. 

\subsection{Training Data}

We train our models on presence-only species observation data obtained from the community science platform iNaturalist~\citep{iNatWeb}. 
The training set consists of 35.5 million observations covering 47,375 species observed prior to 2022. 
Each species observation includes the geographical coordinate where the species was observed. We only included species in the training set if they had at least 50 observations. 
Some species are far more common than others, and thus the dataset is heavily imbalanced (see Figure~\ref{fig:obs_stats}).
Later we use this data in its entirety during training (``All''), with different maximum observations per class (``X / Class''), or with different subsets of classes. 
See Appendix~\ref{sec:sup_train_data} for more details on the training dataset. 

\begin{table*}[ht]
\centering
\caption{
    Results for four geospatial tasks: \textbf{S\&T} (eBird Status \& Trends species mapping), \textbf{IUCN} (IUCN species mapping), \textbf{Geo Prior} (fine-grained image classification with a geographical prior), and \textbf{Geo Feature} (geographical feature regression). Tasks and metrics are defined in Section~\ref{sec:tasks_and_metrics}.We assess performance as a function of the loss function and the amount of training data (``\# / Class"). Model inputs may be coordinates (``Coords."), environmental features (``Env.") or both (``Env. + Coords."). The logistic regression (``LR") and ``Best Discretized Grid" baselines do not have an entry for the \textbf{Geo Feature} task as they do not learn a location encoder. We also do not evaluate models tagged with ``Env." on the \textbf{Geo Feature} task because they are trained on closely related environmental features. Higher values are better for all tasks. 
}
\vspace{3pt}
\label{tab:loss_benchmark}
 \resizebox{1.0\linewidth}{!}{
\begin{tabular}{|l|l| c | c c c c|}
    \hline 
    \multicolumn{3}{c}{} & \textbf{S\&T} & \textbf{IUCN} & \textbf{Geo Prior} & \multicolumn{1}{c}{\textbf{Geo Feature}} \\
    \multicolumn{1}{l}{Loss} & \multicolumn{1}{l}{Model Type} & \multicolumn{1}{l}{\# / Class} & (MAP) & (MAP) & ($\Delta$ Top-1) & \multicolumn{1}{c}{(Mean $R^2$)} \\ \hline
    \multicolumn{6}{l}{\emph{Baselines:}}\\ \hline 
    N/A & Best Discretized Grid~\citep{berg2014birdsnap} & All & 61.56 & 37.13 & +4.1 & -\\ \hline
    $\mathcal{L}_\mathrm{AN-full}$ & LR~\citep{pearce2000evaluation} - Coords. & 1000 & 26.41 & 0.93 & -0.6 & - \\
    $\mathcal{L}_\mathrm{AN-full}$ & LR~\citep{pearce2000evaluation} - Env. &  1000 & 32.91 & 1.23 & -5.6 & -\\ %-10.4
    $\mathcal{L}_\mathrm{AN-full}$ & LR~\citep{pearce2000evaluation} - Env. + Coords. & 1000 & 35.42 & 1.11 & -3.9 & -\\ 
    \hline
    $\mathcal{L}_\mathrm{ME-SSDL}$~\citep{zhou2022acknowledging} & SINR - Coords. & 1000 & 62.74 & 42.55 & +1.6 & 0.726 \\
    $\mathcal{L}_\mathrm{ME-SLDS}$~\citep{zhou2022acknowledging} & SINR - Coords. & 1000 & 74.37 & 32.22 & +2.1 & 0.734 \\
    $\mathcal{L}_\mathrm{ME-full}$~\citep{zhou2022acknowledging} & SINR - Coords. & 1000 & 73.61 & 58.60 & +1.5 & 0.749 \\ \hline 
    $\mathcal{L}_\mathrm{GP}$~\citep{mac2019presence} & SINR - Coords. & 1000 & 73.14 & 59.51 & +5.2 & 0.724 \\ 
    \hline \addlinespace[5pt] \hline
    $\mathcal{L}_\mathrm{AN-SSDL}$ & SINR - Coords. & 10 & 51.12 & 27.63 & +3.4 & 0.631 \\
    $\mathcal{L}_\mathrm{AN-SSDL}$ & SINR - Coords. & 100 & 63.98 & 47.42 & +4.7 & 0.721 \\
    $\mathcal{L}_\mathrm{AN-SSDL}$ & SINR - Coords. & 1000 & 66.99 & 53.47 & +4.9 & 0.744 \\
    $\mathcal{L}_\mathrm{AN-SSDL}$ & SINR - Coords. & All & 68.36 & 55.75 & +4.8 & 0.739 \\
    \hline
    $\mathcal{L}_\mathrm{AN-SLDS}$ & SINR - Coords. & 10 & 63.73 & 27.14 & +4.6 & 0.693 \\
    $\mathcal{L}_\mathrm{AN-SLDS}$ & SINR - Coords. & 100 & 72.18 & 38.40 & +6.1 & 0.731 \\
    $\mathcal{L}_\mathrm{AN-SLDS}$ & SINR - Coords. & 1000 & 76.19 & 42.26 & +6.2 & 0.739 \\
    $\mathcal{L}_\mathrm{AN-SLDS}$ & SINR - Coords. & All & 75.78 & 41.11 & +6.1 & 0.748 \\
    \hline
    $\mathcal{L}_\mathrm{AN-full}$ & SINR - Coords. & 10 & 65.36 & 49.02 & +4.3 & 0.712 \\
    $\mathcal{L}_\mathrm{AN-full}$ & SINR - Coords. & 100 & 72.82 & 62.00 & +6.6 & 0.736 \\
    $\mathcal{L}_\mathrm{AN-full}$ & SINR - Coords. & 1000 & 77.15 & 65.84 & +6.1 & 0.755 \\
    $\mathcal{L}_\mathrm{AN-full}$ & SINR - Coords. & All & 77.94 & 65.59 & +5.0 & 0.759 \\ 
    \hline
    $\mathcal{L}_\mathrm{AN-full}$ & SINR - Env. & 10 & 60.10 & 41.68 & +3.8 & - \\ % 
    $\mathcal{L}_\mathrm{AN-full}$ & SINR - Env. & 100 & 74.54 & 66.64 & +6.7 & - \\ %
    $\mathcal{L}_\mathrm{AN-full}$ & SINR - Env. & 1000 & 79.65 & 70.54 & +6.4 & - \\ % 
    $\mathcal{L}_\mathrm{AN-full}$ & SINR - Env. & All & 80.54 & 69.25 & +5.3 & - \\ \hline  %
    $\mathcal{L}_\mathrm{AN-full}$ & SINR - Env. + Coords. & 10 & 67.12 & 62.99 & +4.7 & - \\ % 
    $\mathcal{L}_\mathrm{AN-full}$ & SINR - Env. + Coords. & 100 & 76.88 & 74.49 & +6.8 & - \\ % 
    $\mathcal{L}_\mathrm{AN-full}$ & SINR - Env. + Coords. & 1000 & 80.48 & 76.07 & +6.5 & - \\ % 
    $\mathcal{L}_\mathrm{AN-full}$ & SINR - Env. + Coords. & All & 81.39 & 74.67 & +5.5 & - \\ % 
    \hline
\end{tabular}
 }
\end{table*}

\subsection{Evaluation Tasks and Metrics}
\label{sec:tasks_and_metrics}

We propose four tasks for evaluating large-scale species range estimation models. We give brief descriptions here, and provide further details in Appendix~\ref{sec:sup_eval_data}.

\noindent\textbf{S\&T: eBird Status and Trends.} 
This task quantifies the agreement between our presence-only predictions and expert-derived range maps from the \textit{eBird Status \& Trends} dataset~\citep{fink2020ebird}, covering 535 bird species with a focus on North America. The spatial extent of this task is visualized in Figure~\ref{fig:snt_iucn_dist}.
Performance is measured using mean average precision (MAP), \ie computing the per-species average precision (AP) and averaging across species. 

\noindent\textbf{IUCN: Expert Range Maps.} 
This task compares our predictions against expert range maps from the International Union for Conservation of Nature (IUCN) Red List~\citep{IUCNRedListData}. 
Unlike the bird-centric \emph{S\&T}, this task covers 2,418 species from different taxonomic groups, including birds, from all over the world. The spatial extent of this task is visualized in Figure~\ref{fig:snt_iucn_dist}. Performance is measured using MAP.

\noindent\textbf{Geo Prior: Geographical Priors for Image Classification.} 
This task measures the utility of our range maps as priors for fine-grained image classification~\citep{berg2014birdsnap,mac2019presence}. 
As illustrated in Figure~\ref{fig:overview}, we combine the output of an image classifier with a range estimation model and measure the improvement in classification accuracy. 
The intuition is that an accurate range model can downweight the probability of a species if it is not typically found at the location where the image was taken. 
For this task we collect 282,974 images from iNaturalist, covering 39,444 species from our training set. 
Each image is accompanied by the latitude and longitude at which the image was taken. 
The performance metric for this task (``$\Delta$ Top-1") is the change in image classifier top-1 accuracy when using our range predictions as a geographical prior. Note that the geographical prior is applied to the classifier at test time -- the image classifier is not trained with any geographical information. A positive value indicates that the prior improves classifier performance. Unlike \emph{S\&T} and \emph{IUCN}, this is an \emph{indirect} evaluation of range map quality since we assess how useful the range predictions are for a downstream task. 

\noindent\textbf{Geo Feature: Environmental Representation Learning.} 
Instead of evaluating the species predictions, this transfer learning task evaluates the quality of the underlying geospatial representation learned by a SINR. The task is to predict nine different geospatial characteristics of the environment, \eg above-ground carbon, elevation, etc. First, we use the location encoder $f_\theta$ to extract features for a grid of evenly spaced locations across the contiguous United States. After splitting the locations into train and test data, we use ridge regression to predict the geospatial characteristics from the extracted features. Performance is evaluated using the coefficient of determination $R^2$ on the test set, averaged across the nine geospatial characteristics. 

\subsection{Results}

\textbf{Which loss is best?} No loss is best in every setting we consider. However, some losses do tend to perform better than others. In Table~\ref{tab:loss_benchmark} we observe that, when we control for input type and the amount of training data, $\mathcal{L}_\mathrm{AN-full}$ outperforms $\mathcal{L}_\mathrm{AN-SSDL}$ and $\mathcal{L}_\mathrm{AN-SLDS}$ most of the time. $\mathcal{L}_\mathrm{AN-full}$ has a decisive advantage on the \emph{S\&T} and \emph{IUCN} tasks and a consistent but small advantage on the \emph{Geo Feature} task. Both $\mathcal{L}_\mathrm{AN-full}$ and $\mathcal{L}_\mathrm{AN-SLDS}$ perform well on the \emph{Geo Prior} task, significantly outperforming $\mathcal{L}_\mathrm{AN-SSDL}$. We note that $\mathcal{L}_\mathrm{AN-full}$ is a simplified version of $\mathcal{L}_\mathrm{GP}$ from~\citet{mac2019presence}, but $\mathcal{L}_\mathrm{AN-full}$ outperforms $\mathcal{L}_\mathrm{GP}$ on every task. 

\textbf{Pseudo-negatives that follow the data distribution are usually better.} $\mathcal{L}_\mathrm{AN-SSDL}$ and $\mathcal{L}_\mathrm{AN-SLDS}$ differ only in the fact that $\mathcal{L}_\mathrm{AN-SSDL}$ samples pseudo-negatives from random locations while $\mathcal{L}_\mathrm{AN-SLDS}$ samples pseudo-negatives from data locations (see Figure~\ref{fig:loss_illustration}). 
In Table~\ref{tab:loss_benchmark} we see that $\mathcal{L}_\mathrm{AN-SLDS}$ outperforms $\mathcal{L}_\mathrm{AN-SSDL}$ for all tasks except \emph{IUCN}. This could be due to the fact that some \emph{IUCN} species have ranges far from areas that are well-sampled by iNaturalist. As we can see in Figure~\ref{fig:three_bird_comparison} (Black Oystercatcher), $\mathcal{L}_\mathrm{AN-SSDL}$ can behave poorly in areas with little training data. This highlights the importance of using diverse tasks to study range estimation methods.

\begin{figure}[tb!]
\centering
\includegraphics[trim={15pt 20pt 10pt 15pt},clip, width=0.4\textwidth]{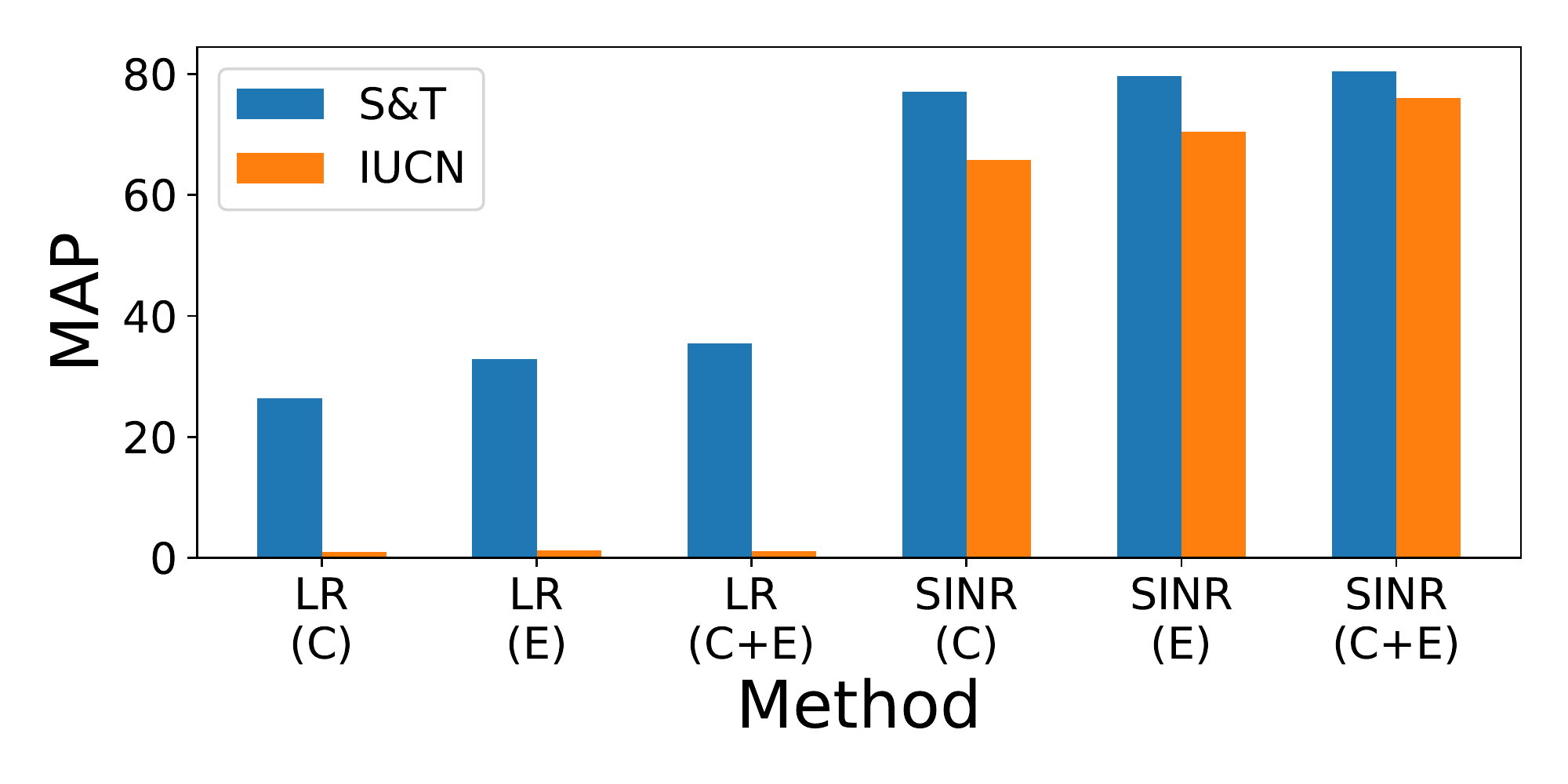}
\vspace{-10pt}
\caption{
    Results for the \emph{S\&T} and \emph{IUCN} tasks. All models are trained with 1000 examples per class using the $\mathcal{L}_\mathrm{AN-full}$ loss. We compare logistic regression (``LR") models against SINR models, using either coordinates (C), environmental covariates (E), or both (C+E) as inputs. These values can also be found in Table~\ref{tab:loss_benchmark}. 
}
\vspace{-15pt}
\label{fig:logistic_regression_comparison}
\end{figure}

\textbf{Implicit neural representations significantly improve performance.} We can assess the impact of the deep location encoder by comparing SINR and LR in models Table~\ref{tab:loss_benchmark}. For instance, if we use the $\mathcal{L}_\mathrm{AN-full}$ loss with 1000 examples per class and coordinates as input, SINR outperforms LR by over 50 MAP on the \emph{S\&T} task. Both methods use the same inputs and training loss -- the only difference is that SINR uses a deep location encoder while LR does not. Figure~\ref{fig:logistic_regression_comparison} shows that same pattern holds whether we use coordinates, environmental features, or both as inputs. For each input type, a deep location encoder provides significant benefits. 

\textbf{Environmental features are not necessary for good performance.} In Figure~\ref{fig:logistic_regression_comparison} we show the \emph{S\&T} and \emph{IUCN} performance of different models trained with coordinates only, environmental features only, or both. We see that SINR models trained with coordinates perform nearly as well as SINR models trained with environmental features. For the SINR models in Figure~\ref{fig:logistic_regression_comparison}, coordinates are 97\% as good as environmental features for the \emph{S\&T} task, 93\% as good for the \emph{IUCN} task, and 95\% as good for the \emph{Geo Prior} task. This suggests that SINRs can successfully use sparse presence-only data to learn about the environment, so that using environmental features as input provides only a marginal benefit. 

\textbf{Coordinates and environmental features are complementary.} Figure~\ref{fig:logistic_regression_comparison} shows that it is better to use the concatenation of coordinates and environmental features than it is to use either coordinates or environmental features alone. This is true for LR and SINR. This indicates that the coordinates and environmental features are carrying some complementary information. However, as we discuss in Appendix~\ref{app:env_vs_coords}, environmental features introduce an additional layer of complexity compared to models that use only coordinates. 

\begin{figure}[tb!]
  \centering
    \includegraphics[trim={0pt 0pt 0pt 0pt},clip,width=0.4\textwidth]{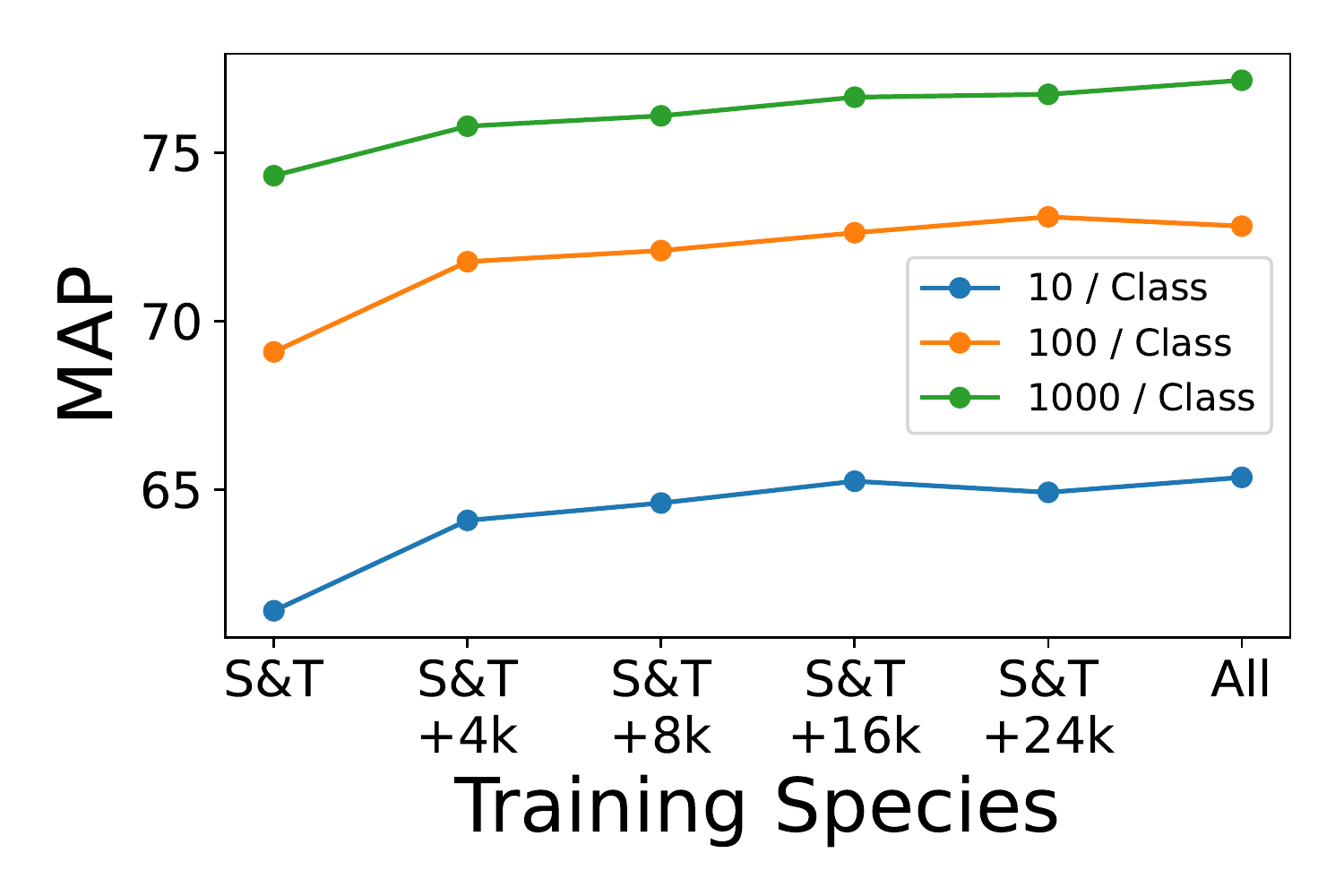}
    \vspace{-15pt}
    \caption{
    \emph{S\&T} task performance with $\mathcal{L}_\mathrm{AN-full}$ as a function of the number of training examples per class (\ie species) and number of classes. The horizontal axis gives the set of species used for training. ``S\&T" indicates that we only train on the 535 species in the S\&T task. For ``S\&T + X" we add in X species chosen uniformly at random. 
    For ``All" we train on all 47k species. 
    Note that the ``10 / Class" point for ``S\&T" is trained with a higher learning rate than usual ($5e-3$ instead of $5e-4$) due to the small number of training examples per epoch. The values for ``All" are also present in Table~\ref{tab:loss_benchmark}. All models use coordinates as input. 
    } \label{fig:num_species}
     \vspace{-15pt}
\end{figure}

\textbf{Joint learning across categories is beneficial, but more data is better.} In Figure~\ref{fig:num_species} we study the effect of the amount of training data on performance for the \emph{S\&T} task. We first note that, unsurprisingly, increasing the number of training examples per species reliably and significantly improves performance. One possible mechanism for this is suggested by Figure~\ref{fig:geo_ica}, which shows a more spatially detailed representation emerging with more training data. 
More interestingly, Figure~\ref{fig:num_species} also shows that adding training data for additional species (which are not evaluated at test time) improves performance as well. That is, the model can better predict the distributions of the \emph{S\&T} birds by also learning the distributions of other birds, plants, insects, \etc  
Intuitively, it seems reasonable that training on more species could lead to a richer and more useful geospatial representation. However, the direct benefit of additional training data for the species of interest is far larger. If we were given a fixed budget of training examples to allocate among species as we wished, we should prefer to have a larger number of training examples per species (instead of fewer training examples per species, but spread across a greater number of species). 

\begin{figure*}[]
\centering
 \resizebox{1.0\linewidth}{!}{
\begin{tabular}{c|c|c|c}
& $\mathcal{L}_\mathrm{AN-SSDL}$ & $\mathcal{L}_\mathrm{AN-SLDS}$ & $\mathcal{L}_\mathrm{AN-full}$ \\
\hline
\rotatebox{90}{\parbox{3cm}{10 Positives / Class}} &
\includegraphics[width=0.28\textwidth]{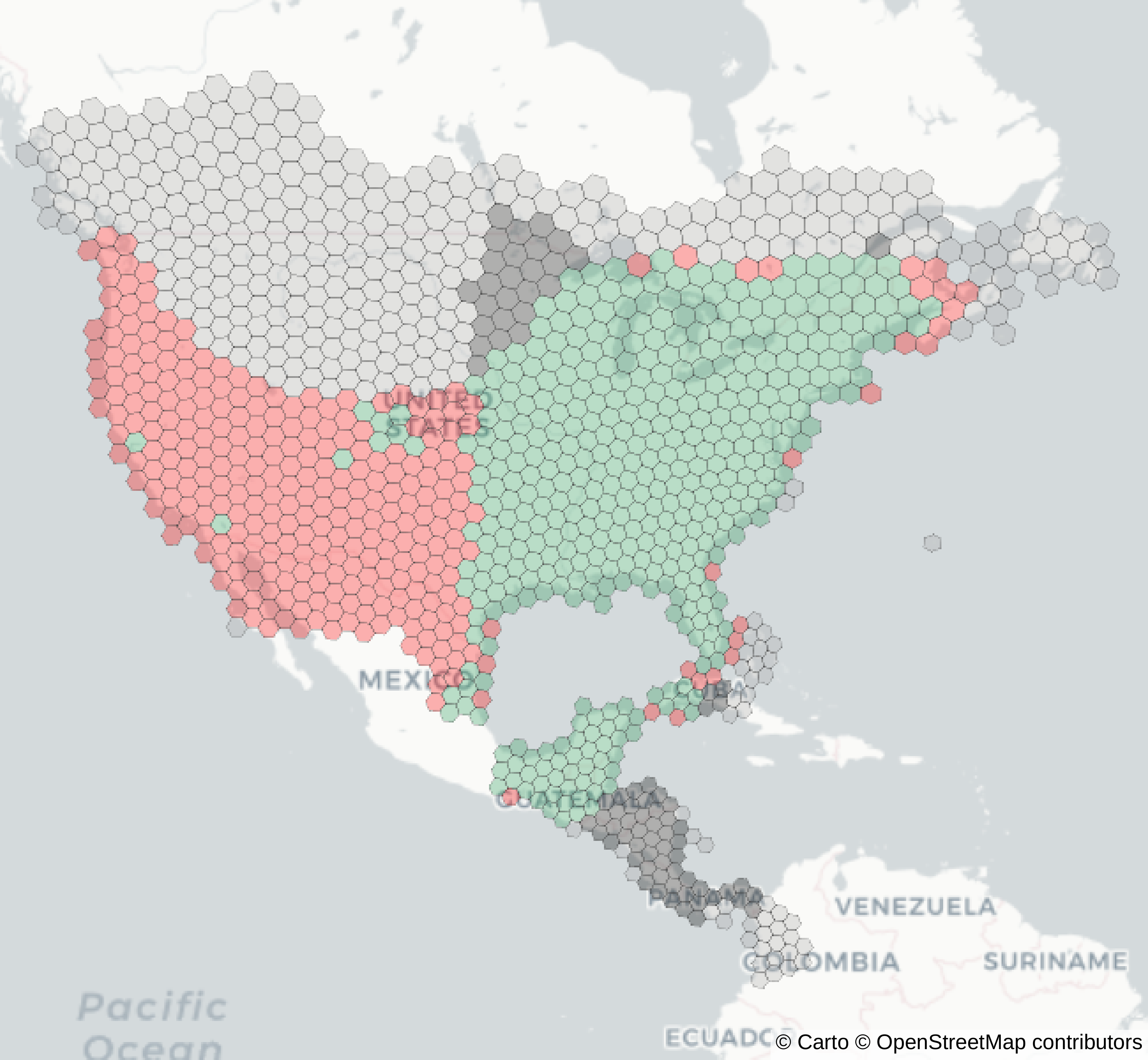} & \includegraphics[width=0.28\textwidth]{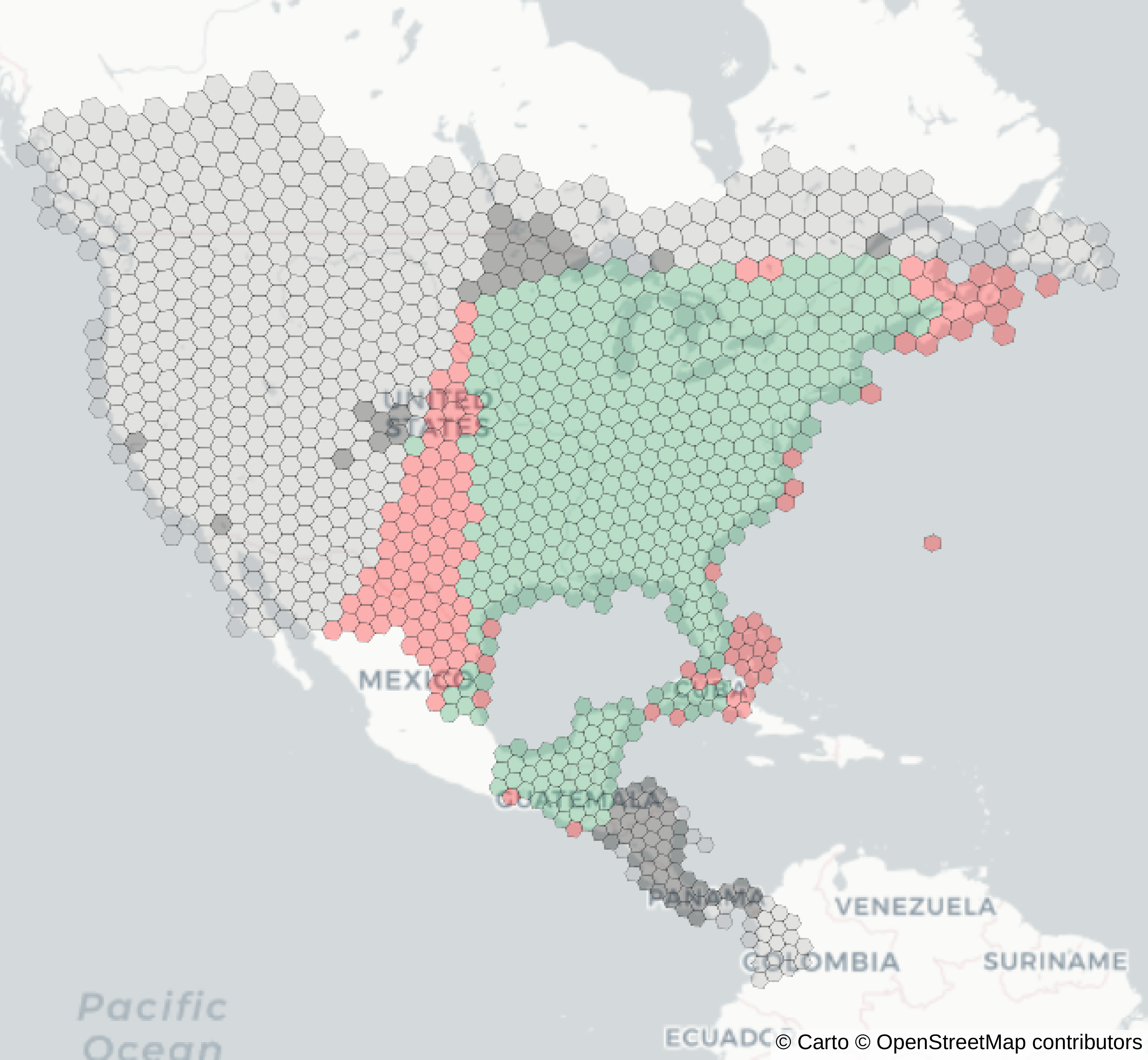} & \includegraphics[width=0.28\textwidth]{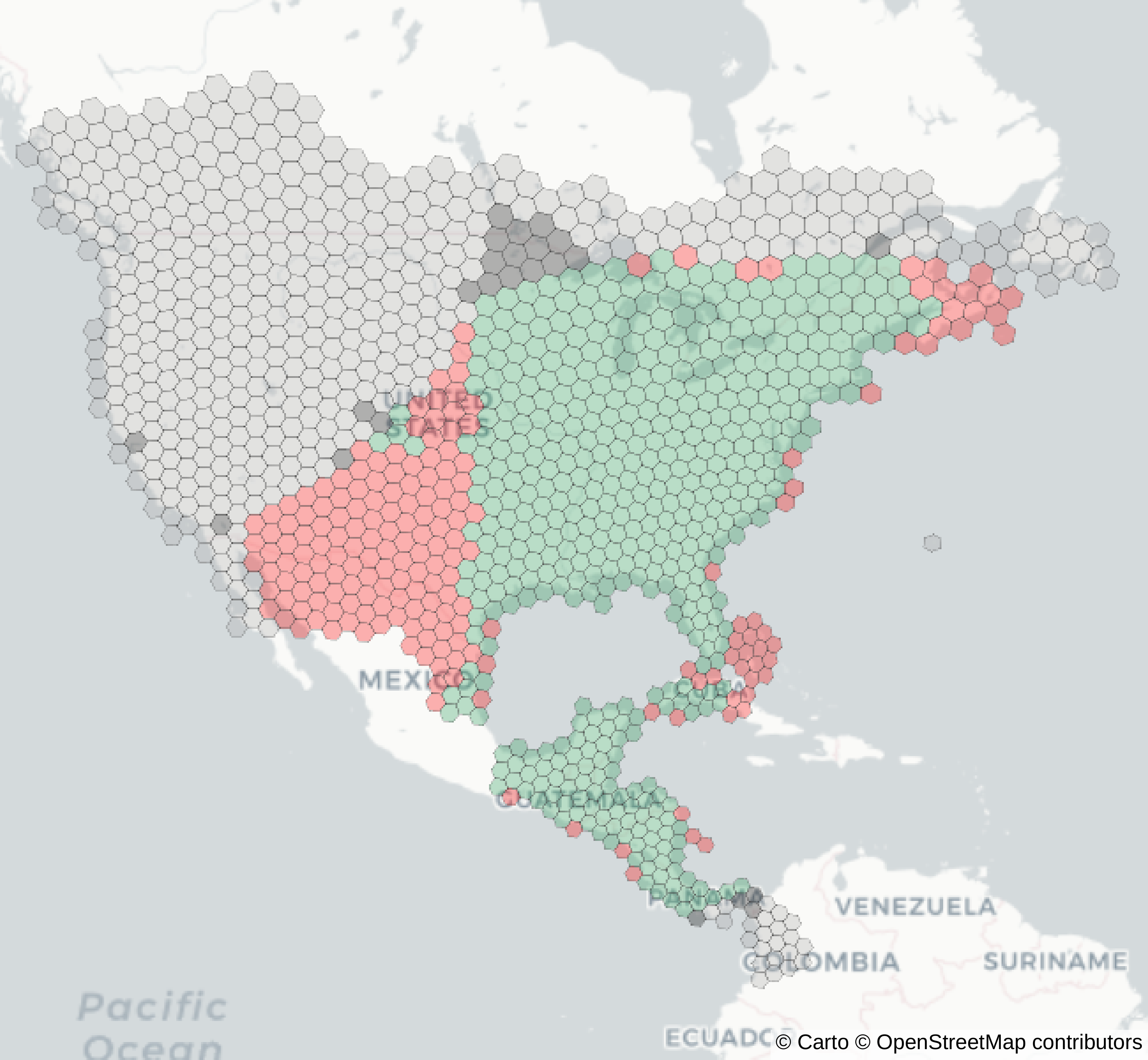} \\
\rotatebox{90}{\parbox{3cm}{100 Positives / Class}} & \includegraphics[width=0.28\textwidth]{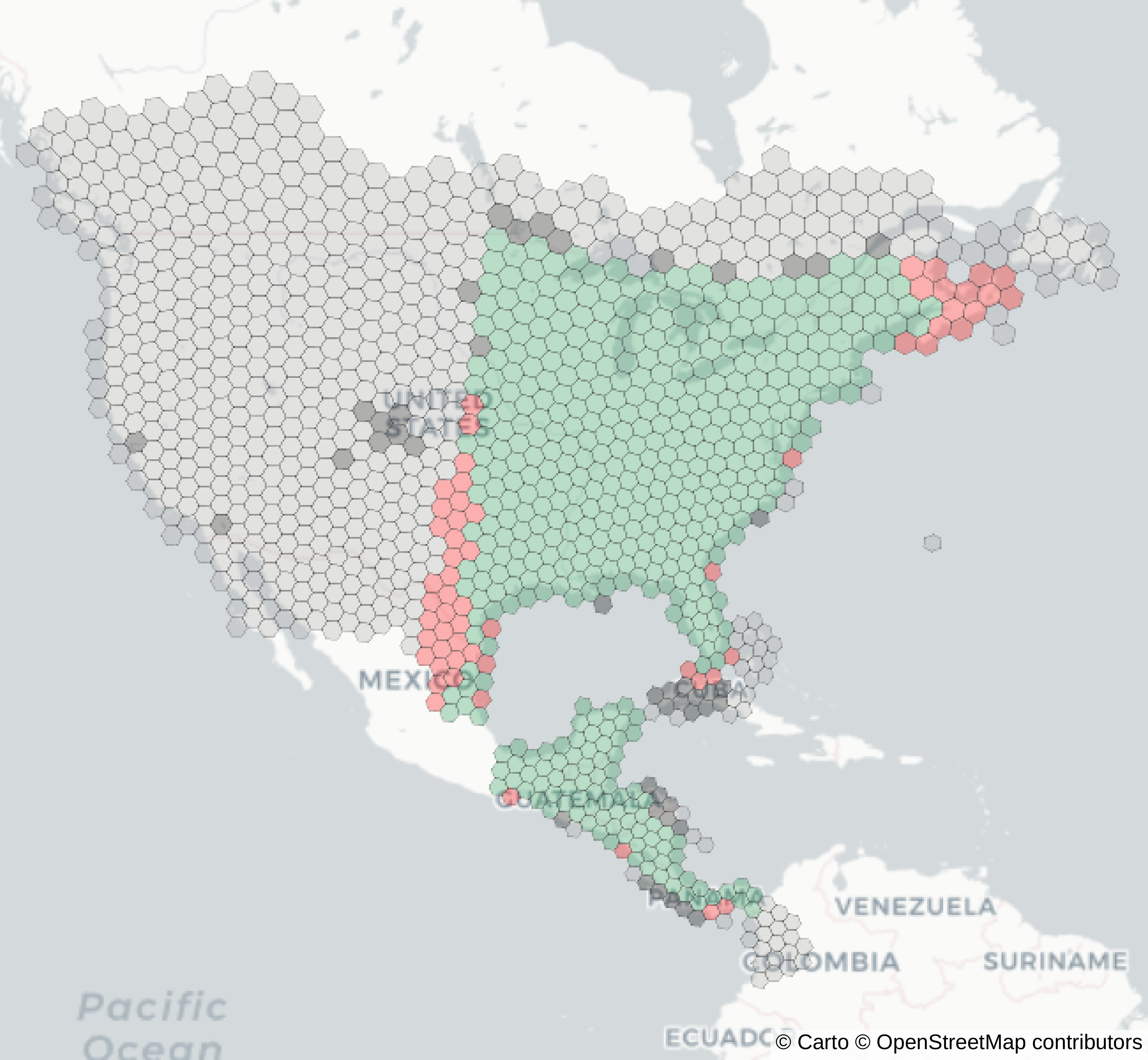} & \includegraphics[width=0.28\textwidth]{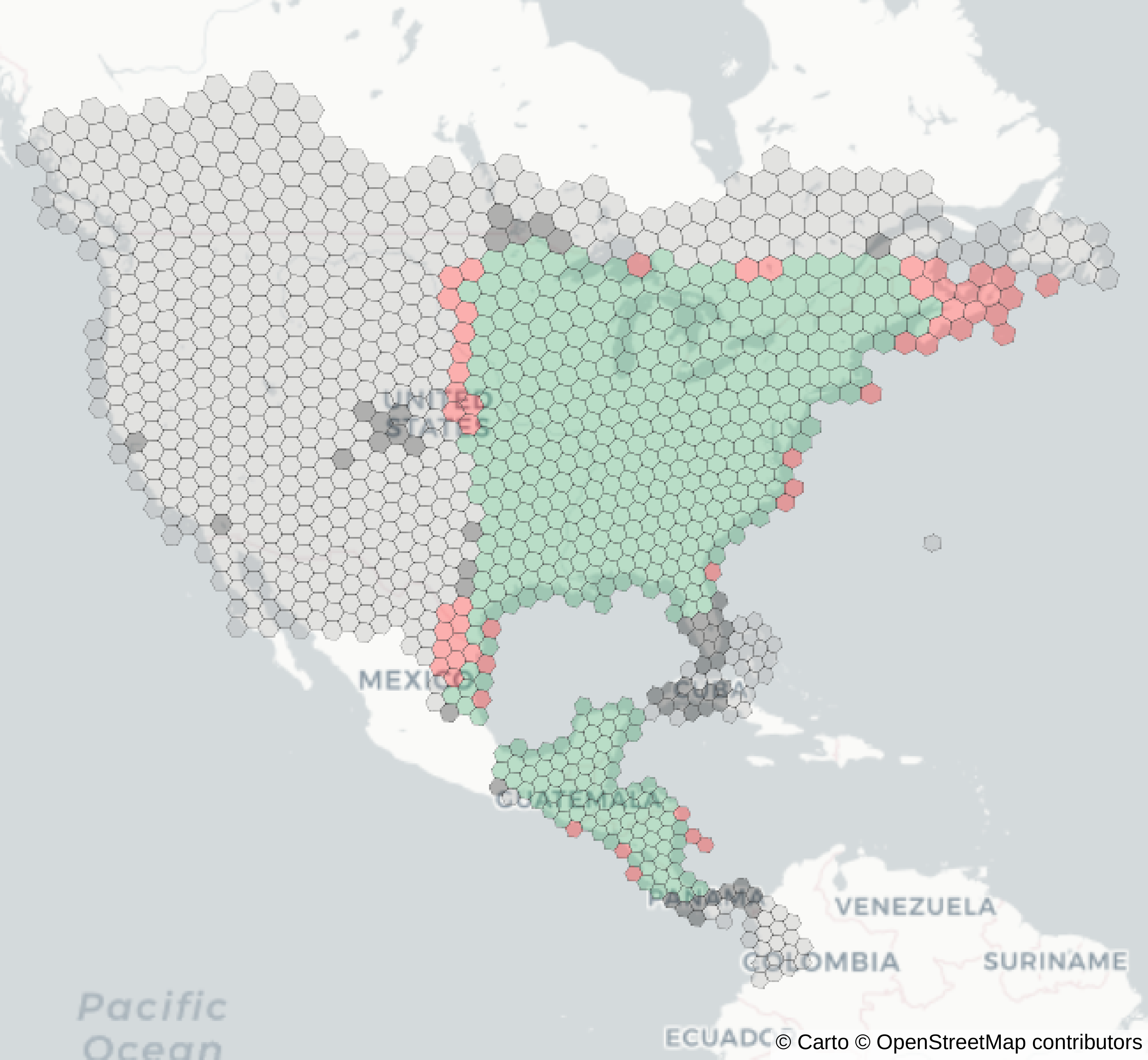} & \includegraphics[width=0.28\textwidth]{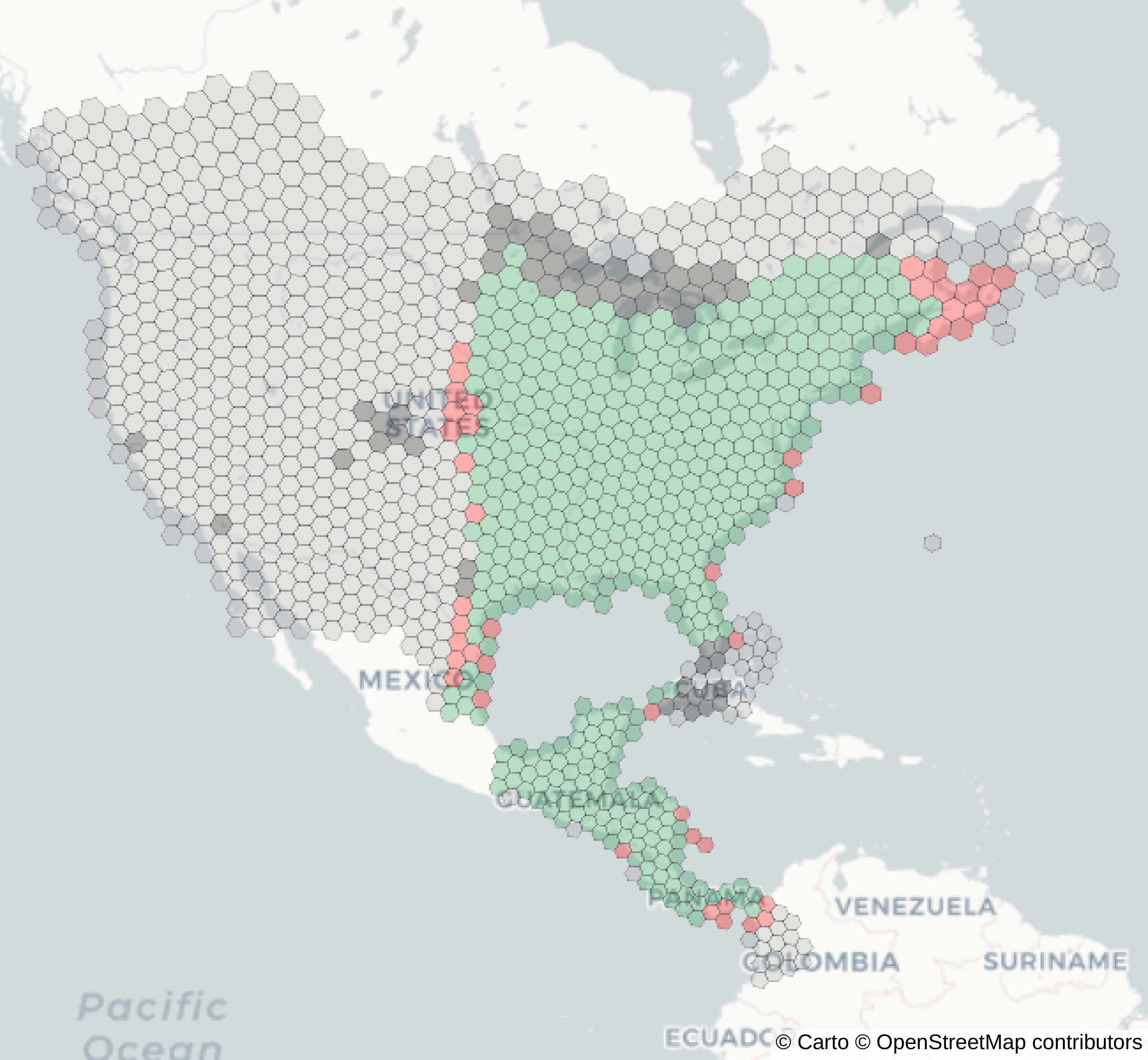} \\
\rotatebox{90}{\parbox{3.2cm}{1000 Positives / Class}} & \includegraphics[width=0.28\textwidth]{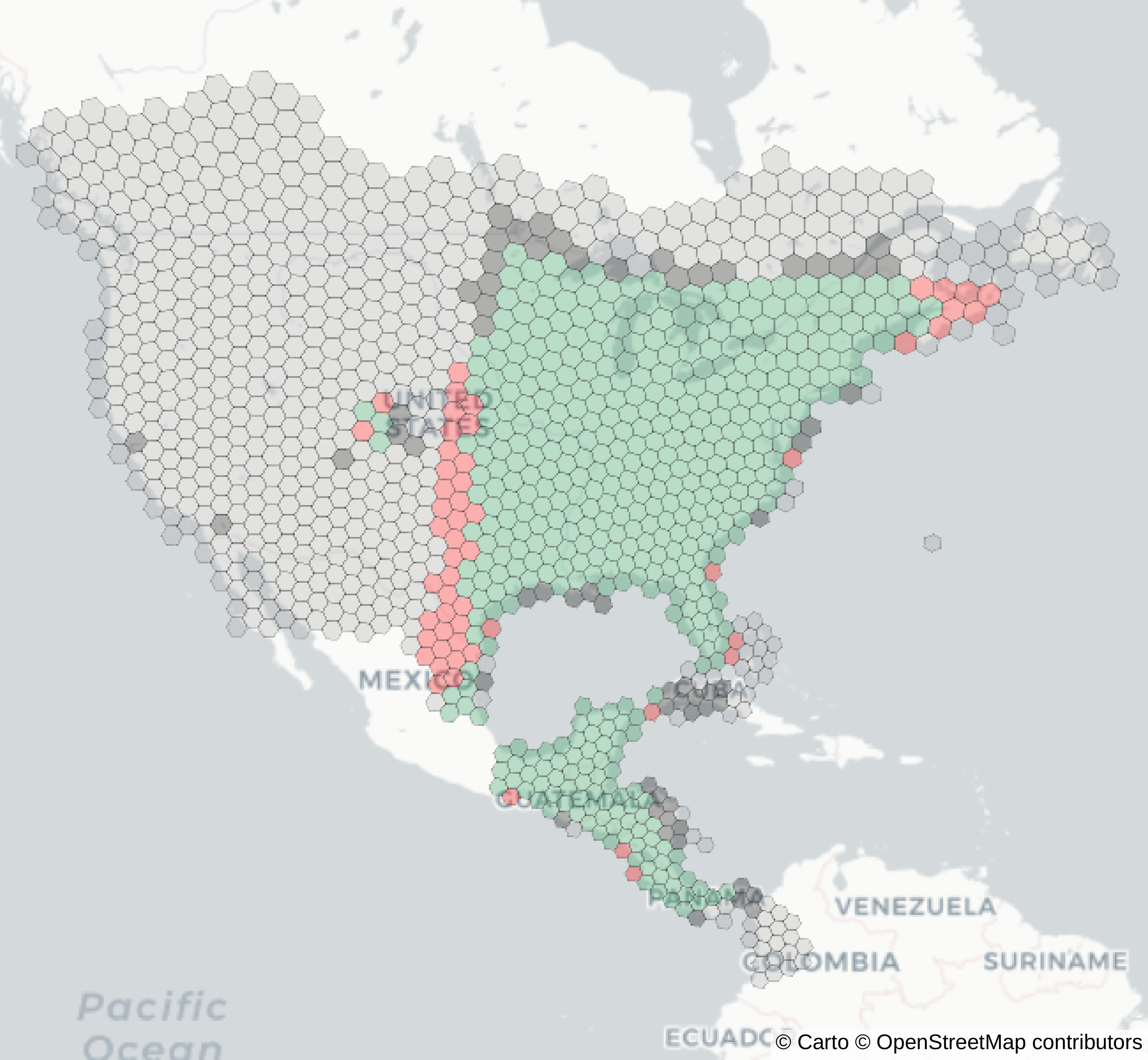} & \includegraphics[width=0.28\textwidth]{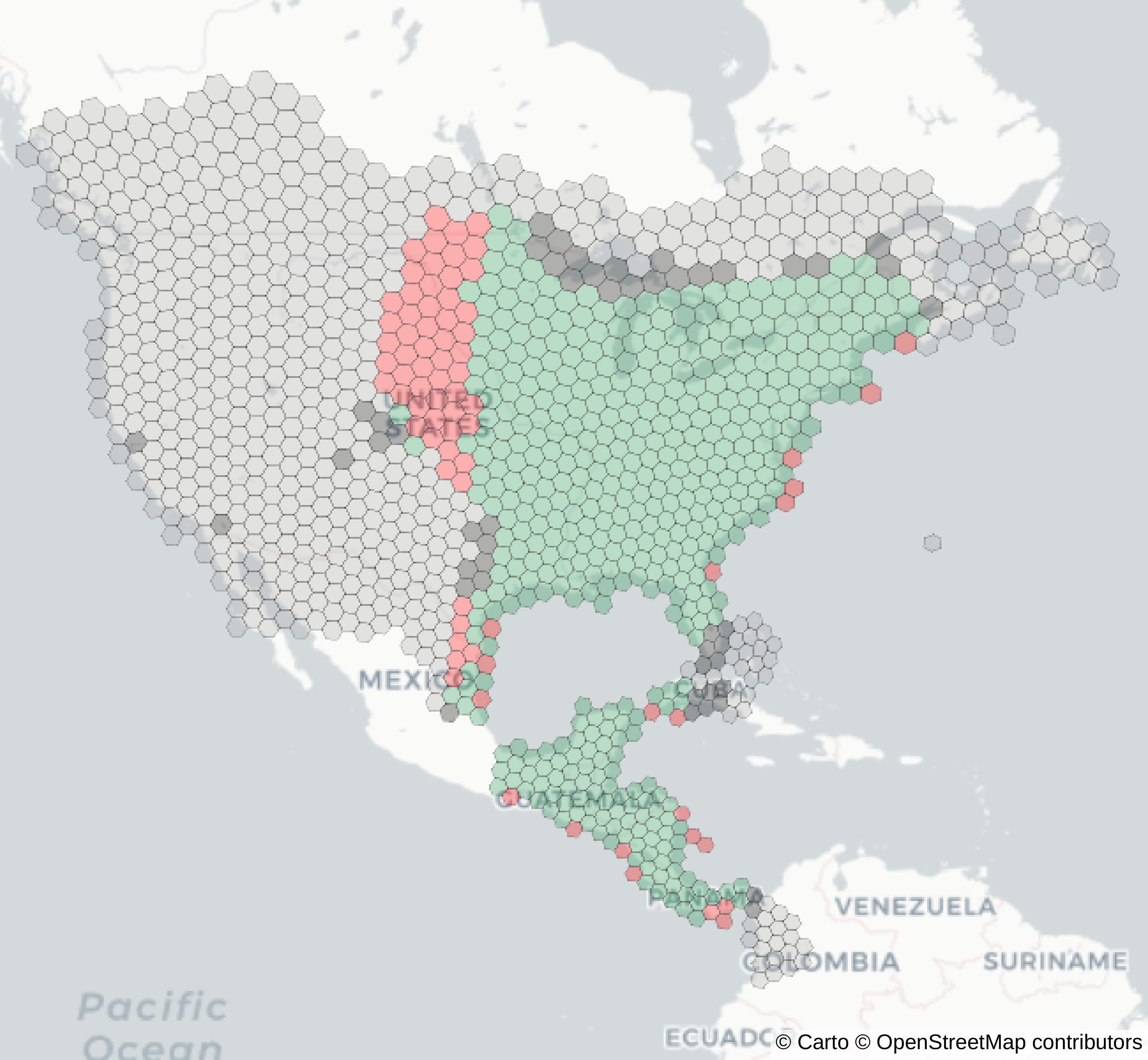} & \includegraphics[width=0.28\textwidth]{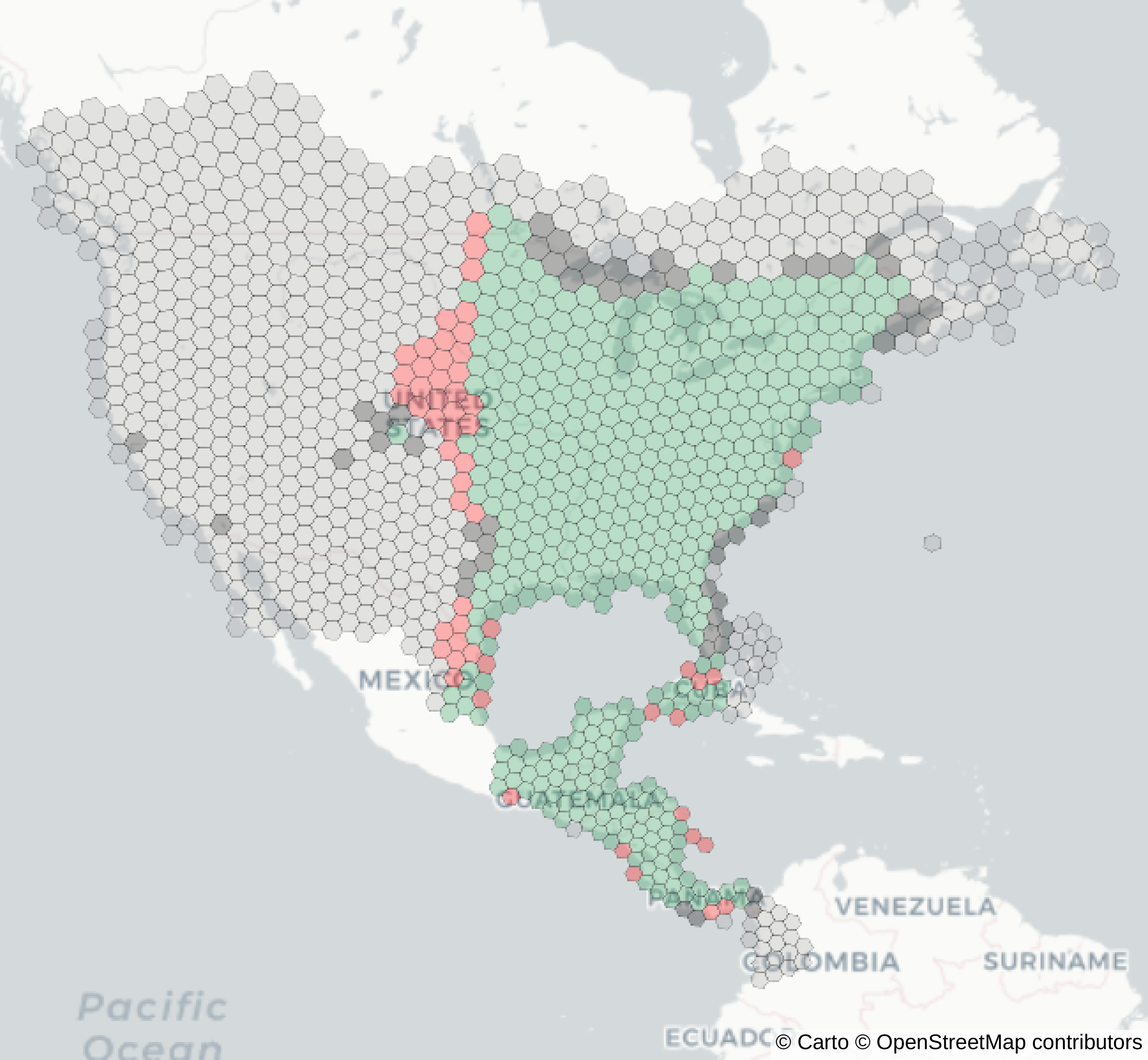} \\
\end{tabular}
 }
\includegraphics[trim={0pt 0pt 0pt 0pt},clip,height=20px]{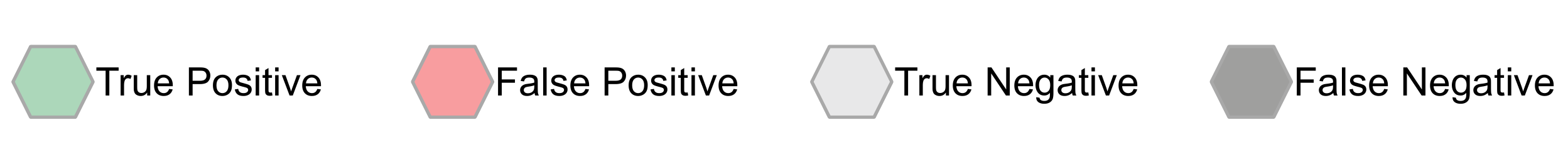}
\hspace{5pt}
\caption{
    Visualization of SINR predictions for \href{https://ebird.org/species/woothr}{Wood Thrush} when varying the amount of training data (rows) for different loss functions (columns). Model predictions are generated at the centroid of the rendered hexagons for a coarse H3 grid (resolution three), signifying locations where we can evaluate the model outputs for the \emph{S\&T} task. We convert the predictions to binary values using the threshold that maximizes the F1 score on the \emph{S\&T} data. This is done for each configuration independently. In practice this threshold would be chosen by a practitioner to meet particular project requirements. A model that matches the \emph{S\&T} task exactly would show only green and light grey hexagons. All models improve their range maps when given access to more data, as expected. $\mathcal{L}_\mathrm{AN-SSDL}$ overestimates the western range extent and misses the southern extent with few examples, but refines these extents with additional data. $\mathcal{L}_\mathrm{AN-full}$ starts off with most of the range covered (few ``False Negative" hexagons) and proceeds to tighten the boundaries with more data. The range predicted by $\mathcal{L}_\mathrm{AN-SLDS}$ is somewhere in between. All models use coordinates as input. 
}
\label{fig:woothr_comparison}
\vspace{-10pt}
\end{figure*}

\textbf{Low-shot performance is surprisingly good.} In Table~\ref{tab:loss_benchmark} we see that a SINR trained with $\mathcal{L}_\mathrm{AN-full}$ and only 10 examples per category (\ie $\sim$1\% of the training data) beats the ``Best Discretized Grid'' baseline (which uses all of the training data) on every task. SINRs seem to be capable of capturing general spatial patterns using relatively little data. While this is encouraging, we expect that more data is necessary to capture fine detail as suggested by Figure~\ref{fig:geo_ica} and Figure~\ref{fig:woothr_comparison}.

\begin{figure}[tb!]
\centering
\includegraphics[trim={30pt 15pt 40pt 35pt},clip, width=0.35\textwidth]{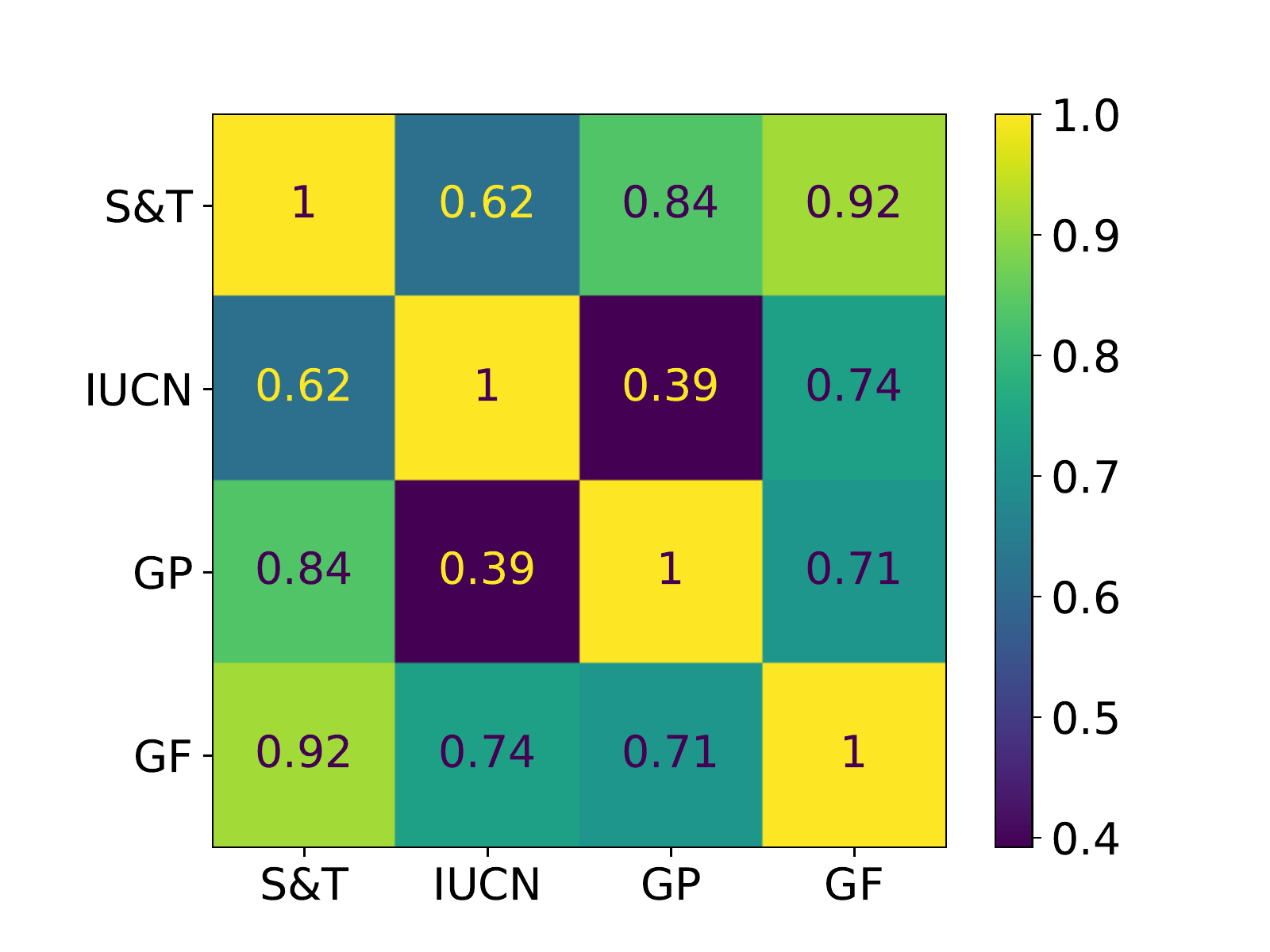}
\vspace{-10pt}
\caption{
    Performance correlations across our four tasks: \emph{S\&T}, \emph{IUCN}, \emph{Geo Prior} (GP), and \emph{Geo Feature} (GF). Values are Pearson product-moment correlation coefficients. The correlations are computed across 12 SINR models: $\mathcal{L}_\mathrm{AN-SSDL}$, $\mathcal{L}_\mathrm{AN-SLDS}$, and $\mathcal{L}_\mathrm{AN-full}$ for 10, 100, 1000, and All training examples per class. All models use coordinates as input. 
}
\vspace{-15pt}
\label{fig:task_correlation}
\end{figure}

\textbf{How are our tasks related?} In this work we study four spatial prediction tasks. This tasks differ in their spatial domains, evaluation metrics, and categories of interest, but it is reasonable to wonder to what extent they may be related. In Figure~\ref{fig:task_correlation} we show the pairwise correlations between scores on our tasks. Some tasks are highly correlated (\eg \emph{S\&T} and \emph{Geo Features}, 0.92) while others are not (\eg \emph{IUCN} and \emph{Geo Prior}, 0.39). 

\textbf{Imbalance hurts performance, but not too much.} In Table~\ref{tab:loss_benchmark} we notice that a SINR trained with all of the training data often performs worse than a SINR trained on up to 1000 examples per class. This pattern is clearest for the \emph{IUCN} and \emph{Geo Prior} tasks. Capping the number of training examples per class reduces the amount of training data, but it also reduces class imbalance in the training set (some categories have as many as $\sim 10^5$ training examples). It seems that the benefit of reducing class imbalance outweighs the benefit of additional training data in these cases. However, it is important to keep in mind that the performance drops we are discussing are small. For instance, for a SINR trained with  $\mathcal{L}_\mathrm{AN-full}$ and coordinates as input, switching from 1000 training examples to all of the training data changes performance by -0.79 MAP for the \emph{S\&T} task, -0.25 MAP for the \emph{IUCN} task, -1.1 $\Delta$ Top-1 for the \emph{Geo Prior} task, and +0.004 for the \emph{Geo Feature} task. Given the extreme imbalance in the training set and the fact that we do not explicitly handle class imbalance during training, it may be surprising that the performance drops are not larger.

\textbf{Loss function rankings may not generalize across domains.} The presence-only SDM problem in this work and the single positive image classification problem in \citet{cole2021multi} are both SPML problems. Despite this formal equivalence, it does not seem that the best methods for SPML image classification are also the best methods for presence-only SDM. \citet{zhou2022acknowledging} show that their ``maximum entropy" loss performs much better than the ``assume negative" loss across a number of image classification datasets. However, all of the ``maximum entropy" losses in Table~\ref{tab:loss_benchmark} ($\mathcal{L}_\mathrm{ME-SSDL}$, $\mathcal{L}_\mathrm{ME-SLDS}$, $\mathcal{L}_\mathrm{ME-full}$) underperform their ``assume negative" counterparts ($\mathcal{L}_\mathrm{AN-SSDL}$, $\mathcal{L}_\mathrm{AN-SLDS}$, $\mathcal{L}_\mathrm{AN-full}$). Thus, the benchmarks in this paper are complementary to those in \citet{cole2021multi} and may be useful in developing a more holistic understanding of SPML learning. 

\subsection{Limitations} 
\vspace{-5pt}

It is important to be aware of the limitations associated with our analysis. 
As noted, the training set is heavily imbalanced, both in terms of the species themselves and where the data was collected. 
In practice, some of the most biodiverse regions are underrepresented. 
This is partially because some species are more common and thus more likely to be observed than others by iNaturalist users.  
We do not explicitly deal with species imbalance in the training data, other than by showing that the ranking of methods does not significantly vary even when the training data for each species is capped to the same upper limit (see Table~\ref{tab:loss_benchmark}). 

Reliably evaluating the performance of SDMs for many species and locations is a long standing challenge. 
To address this issue, we present a suite of complementary benchmarks that attempt to evaluate different facets of this spatial prediction problem. 
However, obtaining ground truth range data for thousands of species remains very difficult. 
While we believe our benchmarks to be a significant step forward, they are likely to have blind spots, \eg they are limited to well-described species and can contain inaccuracies.  

Finally, care should be taken before making conservation decisions based on the outputs of models such as the ones presented here. Our goal in this work is to demonstrate the promise of large-scale representation learning for species distribution modeling. Our models have not been calibrated or validated beyond the experiments illustrated above. 

\vspace{-5pt}
\section{Conclusion} 
We explored the problem of species range mapping through the lens of learning spatial implicit neural representations (SINRs). 
In doing so, we connected recent work on implicit coordinate networks and learning multi-label classifiers from limited supervision. 
We hope our contributions encourage more machine learning researchers to work on this important problem. While the initial results are encouraging, there are many avenues for future work. For example, our models make no use of time~\citep{mac2019presence}, do not account for spatial bias~\citep{chen2019bias}, and have no inductive biases for encoding spatially varying signals~\citep{ramasinghe2021beyond}. 

\noindent{\bf Acknowledgments.} We thank the iNaturalist and eBird communities for their data collection efforts, as well as Matt Stimas-Mackey and Sam Heinrich for help with data curation. This project was funded by the Climate Change AI Innovation Grants program, hosted by Climate Change AI with the support of the Quadrature Climate Foundation, Schmidt Futures, and the Canada Hub of Future Earth. 
This work was also supported by the Caltech Resnick Sustainability Institute and an NSF Graduate Research Fellowship (grant number DGE1745301).

%\clearpage
\bibliography{main}
\bibliographystyle{icml2023}

%%%%%%%%%%%%%%%%%%%%%%%%%%%%%%%%%%%%%%%%%%%%%%%%%%%%%%%%%%%%%%%%%%%%%%%%%%%%%%%
%%%%%%%%%%%%%%%%%%%%%%%%%%%%%%%%%%%%%%%%%%%%%%%%%%%%%%%%%%%%%%%%%%%%%%%%%%%%%%%
% APPENDIX
%%%%%%%%%%%%%%%%%%%%%%%%%%%%%%%%%%%%%%%%%%%%%%%%%%%%%%%%%%%%%%%%%%%%%%%%%%%%%%%
%%%%%%%%%%%%%%%%%%%%%%%%%%%%%%%%%%%%%%%%%%%%%%%%%%%%%%%%%%%%%%%%%%%%%%%%%%%%%%%
\newpage
\appendix

\setcounter{table}{0}
\renewcommand{\thetable}{A\arabic{table}}
\setcounter{figure}{0}
\renewcommand{\thefigure}{A\arabic{figure}}

\onecolumn
\input{supp_content.tex}

\end{document}

%% file: supp_content.tex
{\huge Appendix}

\section{Additional Results}
\label{sec:supp_results}

\subsection{How much does performance vary when we re-train a SINR?}\label{sec:supp_variance}

\begin{figure}[htb!]
\centering
\includegraphics[trim={0pt 0pt 0pt 0pt},clip, width=0.5\textwidth]{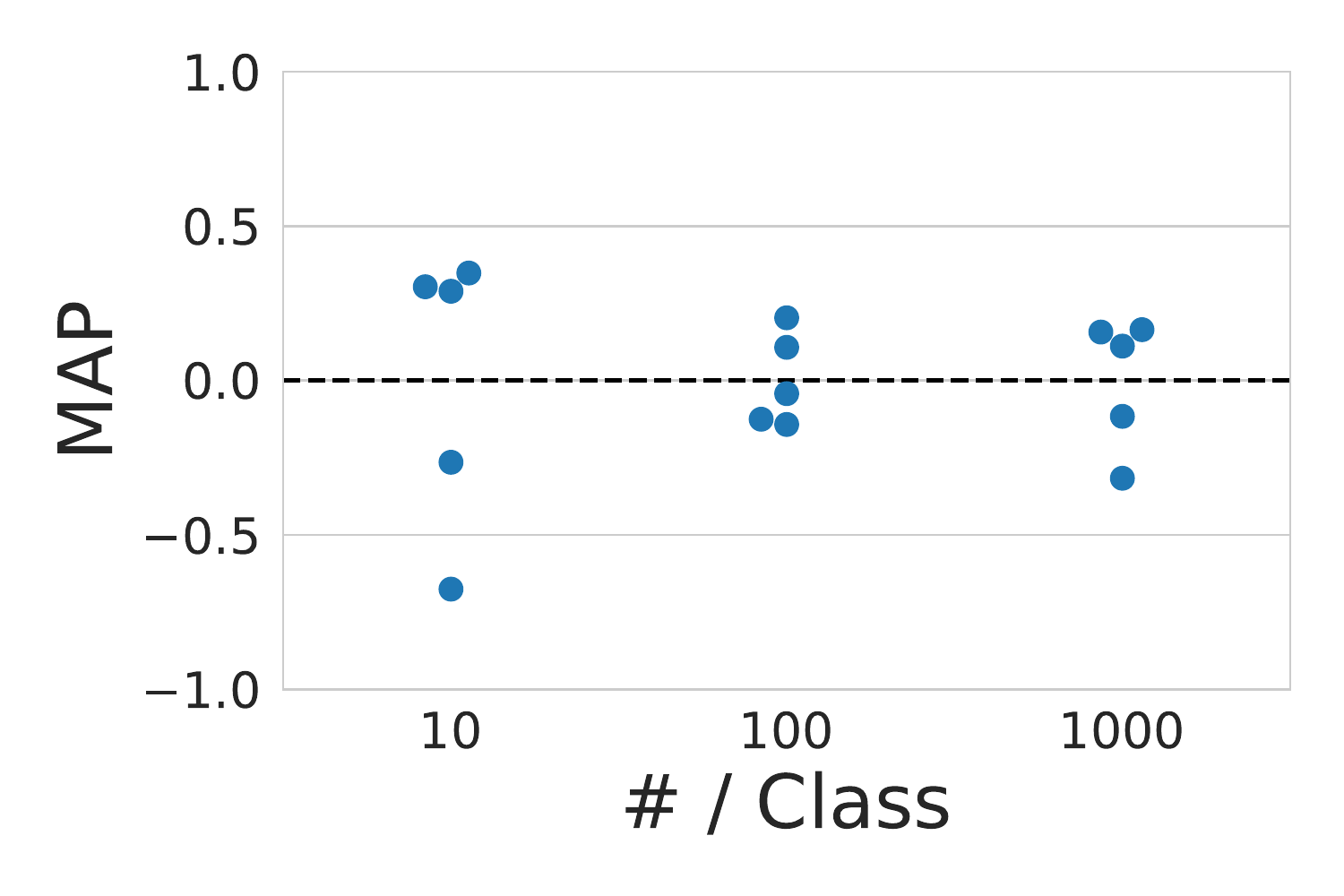}
\vspace{-8pt}
\caption{
    \emph{S\&T} results for SINRs trained with the $\mathcal{L}_\mathrm{AN-full}$ loss and varying amounts of training data. For each training data level, we show the mean-subtracted performance for 5 independent training runs. For this figure, the training examples selected for each class are re-sampled for each run. Thus, the randomness we see in this figure combines the randomness due to retraining and the randomness due to training data selection. Deviation from the mean is typically less the $0.5$ MAP, and is always less than $1.0$ MAP. All models use coordinates as input. 
}
\label{fig:retrain}
\end{figure}

The goal of this section is to provide a sense for how much variance in the performance of a SINR is due to randomness in the training process. We show \emph{S\&T} results for multiple independently trained SINRs in Figure~\ref{fig:retrain}. First, we observe that (as expected) performance varies more when training on 10 examples per class than it does when training on 100 or 1000 examples per class. Second, we note that deviation from the mean is typically less than 0.5 MAP and always less than 1.0 MAP. This provides some context for understanding whether a difference between two models is likely to be ``real" or merely due to randomness. 

\subsection{Additional Qualitative Results}\label{sec:supp_qualitative}

\begin{figure}[h]
\centering
\resizebox{1.0\linewidth}{!}{
\begin{tabular}{c|c|c|c}
& $\mathcal{L}_\mathrm{AN-SSDL}$ & $\mathcal{L}_\mathrm{AN-SLDS}$ & $\mathcal{L}_\mathrm{AN-full}$ \\
\hline
\rotatebox{90}{\parbox{2cm}{\tiny{Barn Swallow}}} &
\includegraphics[width=0.28\textwidth]{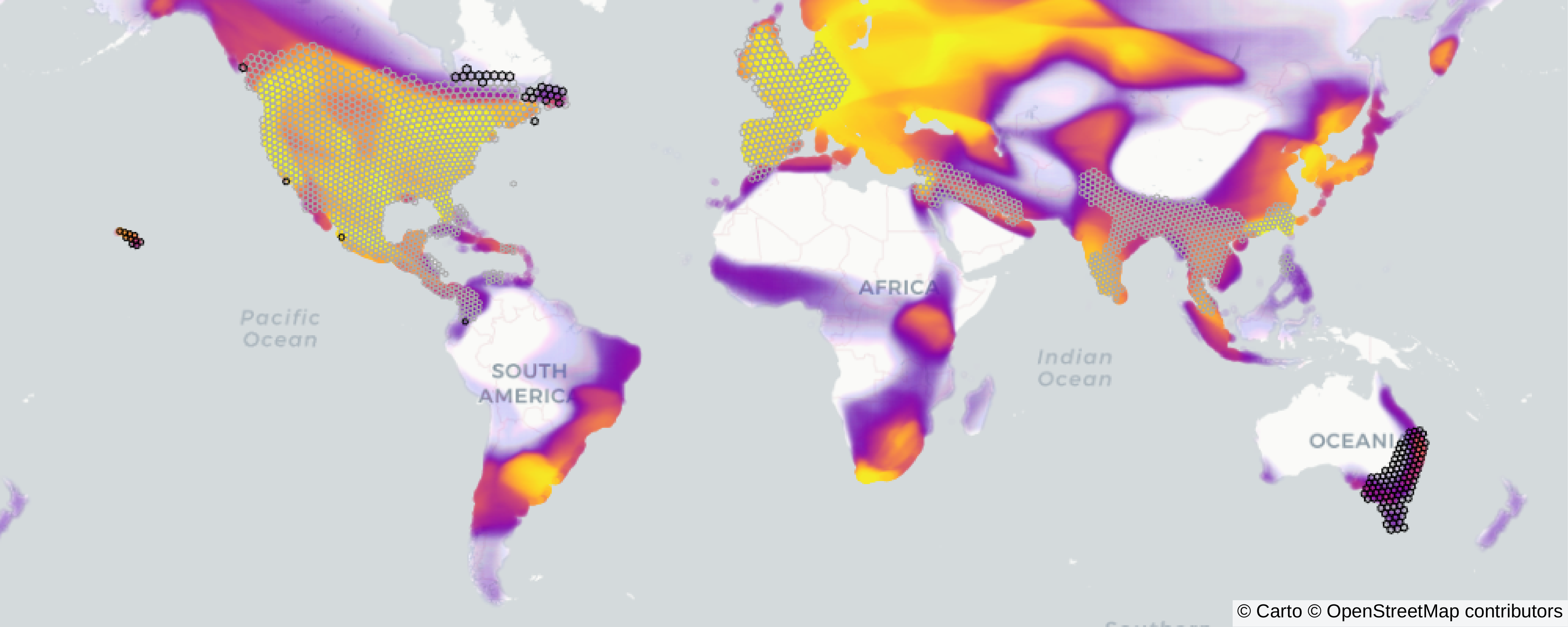} & \includegraphics[width=0.28\textwidth]{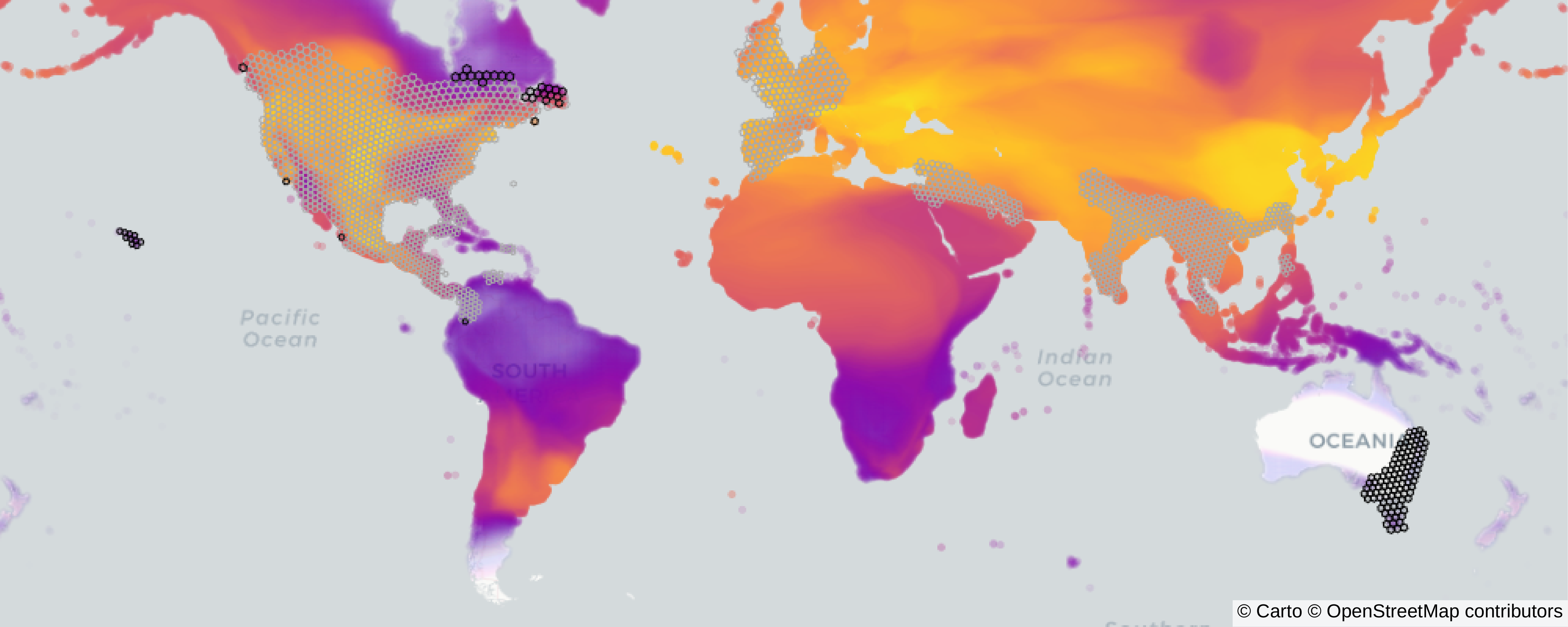} & \includegraphics[width=0.28\textwidth]{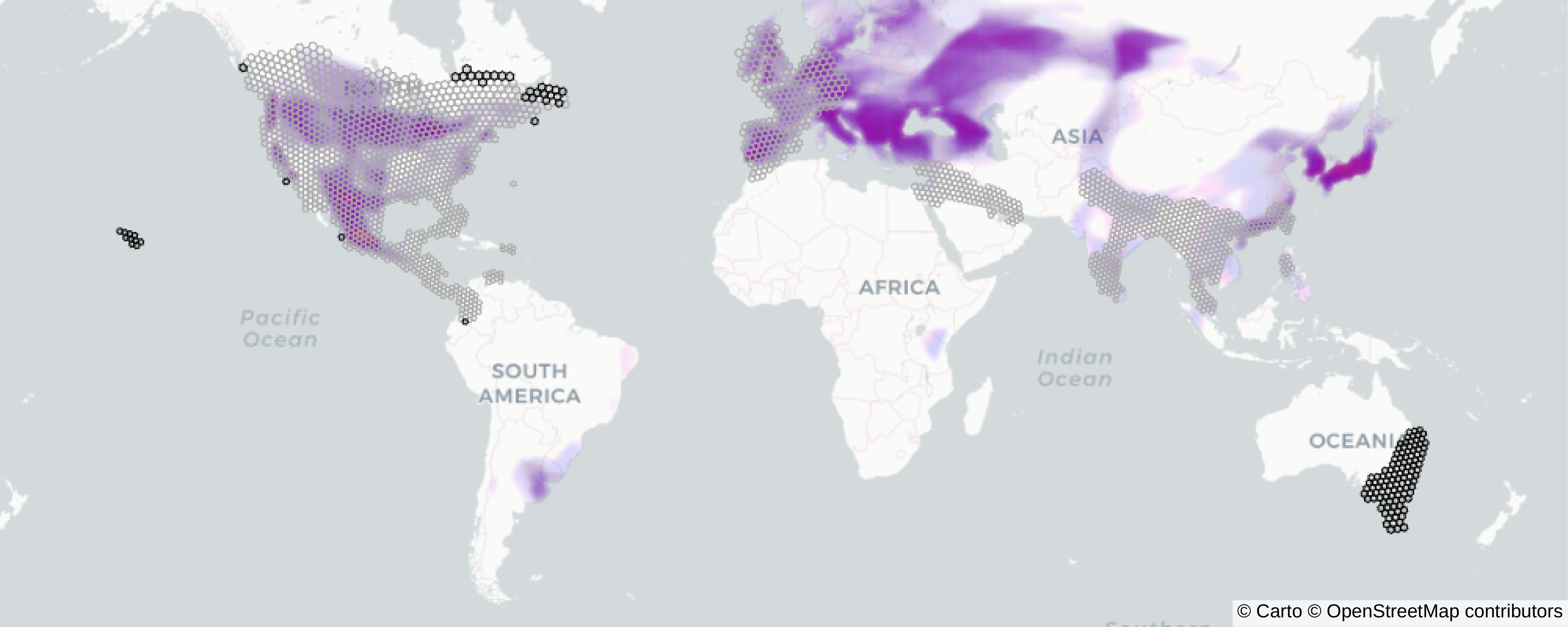} \\
\rotatebox{90}{\parbox{2cm}{\tiny{Blue Rock-Thrush}}} & \includegraphics[width=0.28\textwidth]{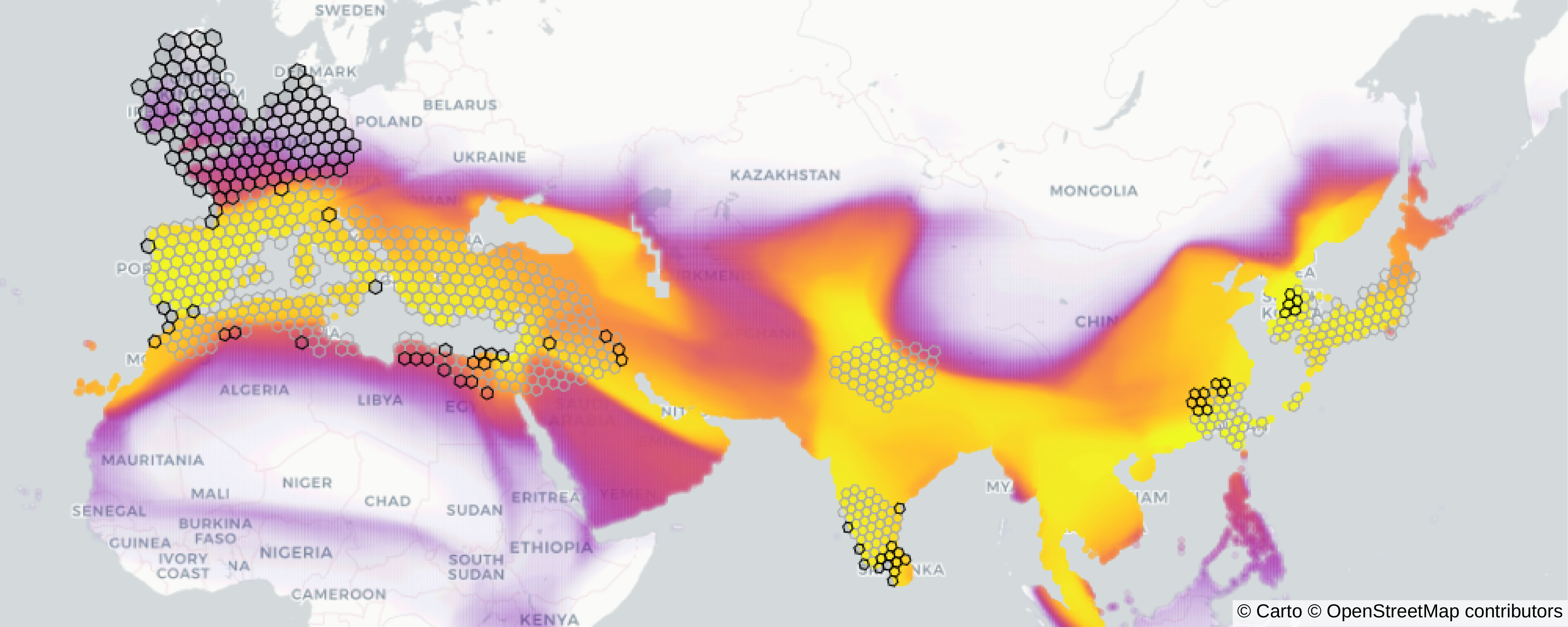} & \includegraphics[width=0.28\textwidth]{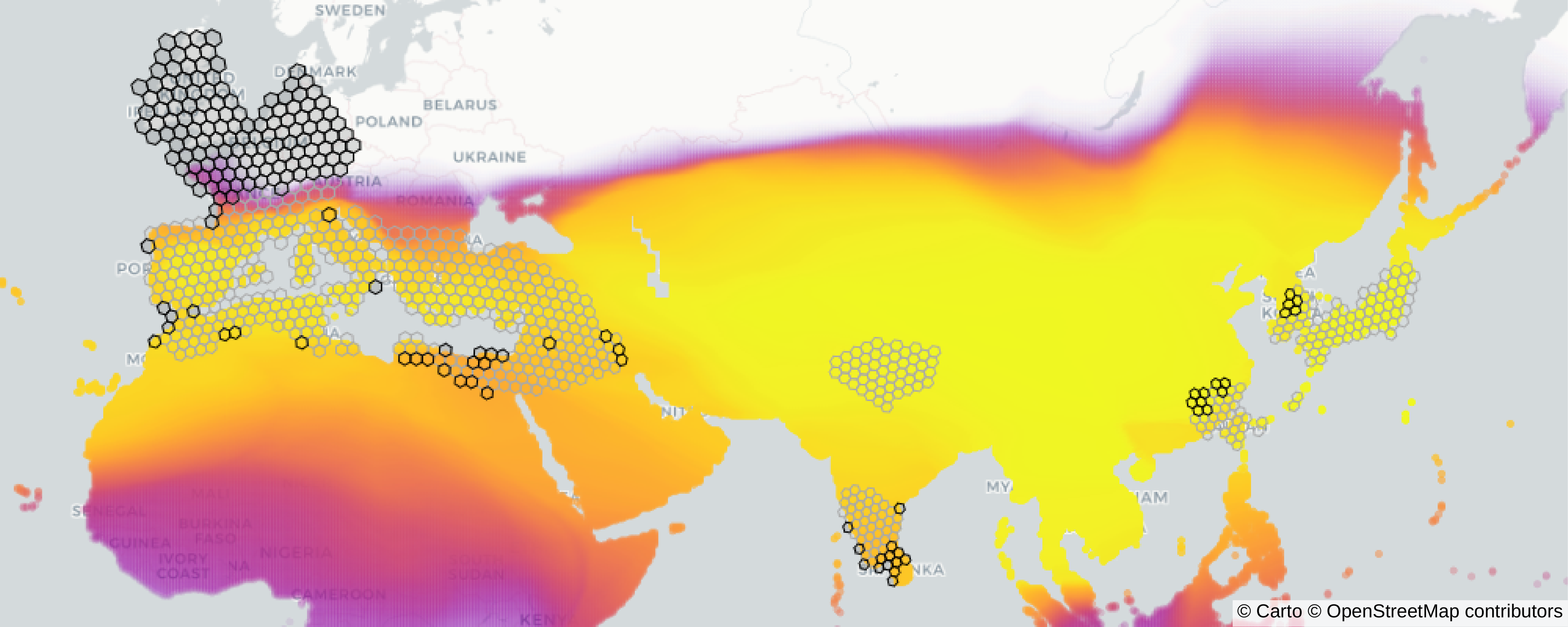} & \includegraphics[width=0.28\textwidth]{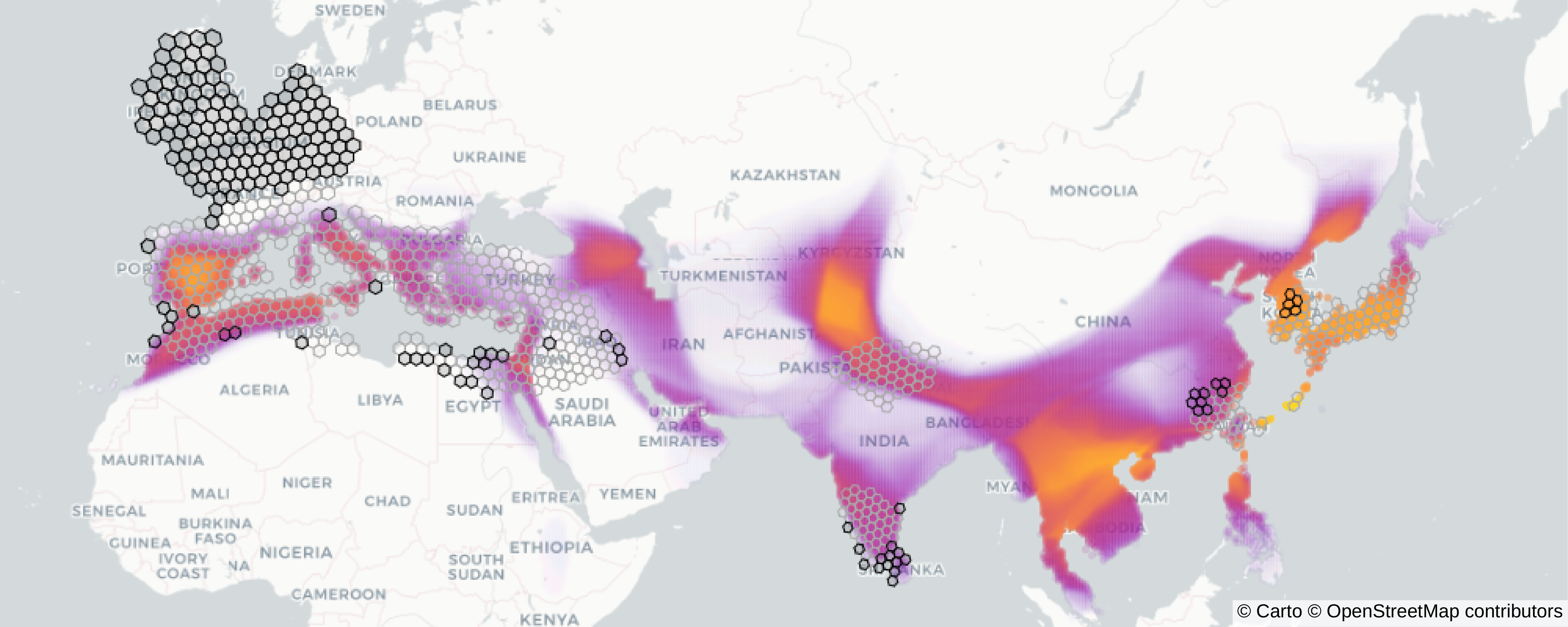} \\
\rotatebox{90}{\parbox{2cm}{\tiny{Black Oystercatcher}}} & \includegraphics[width=0.28\textwidth]{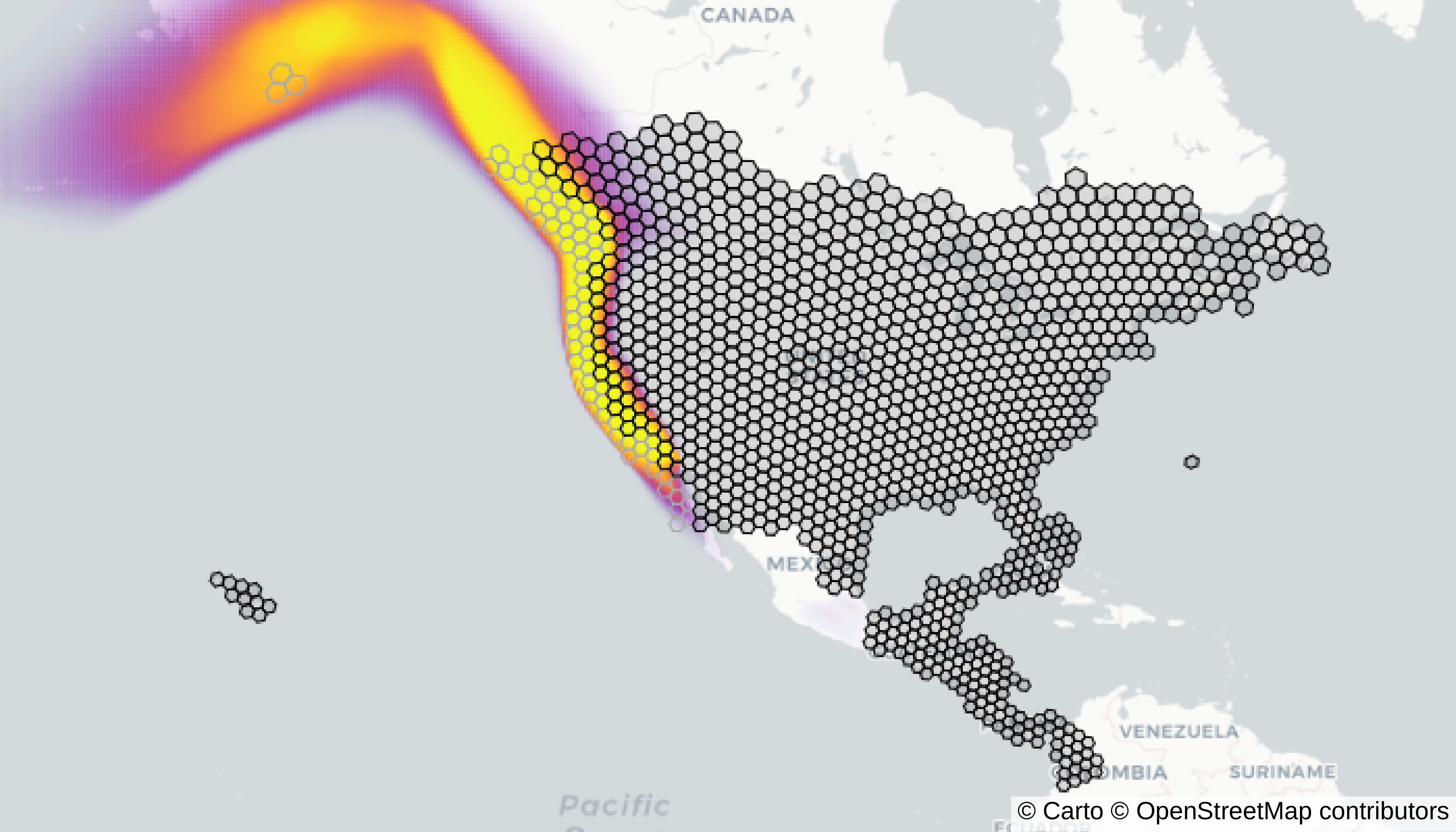} & \includegraphics[width=0.28\textwidth]{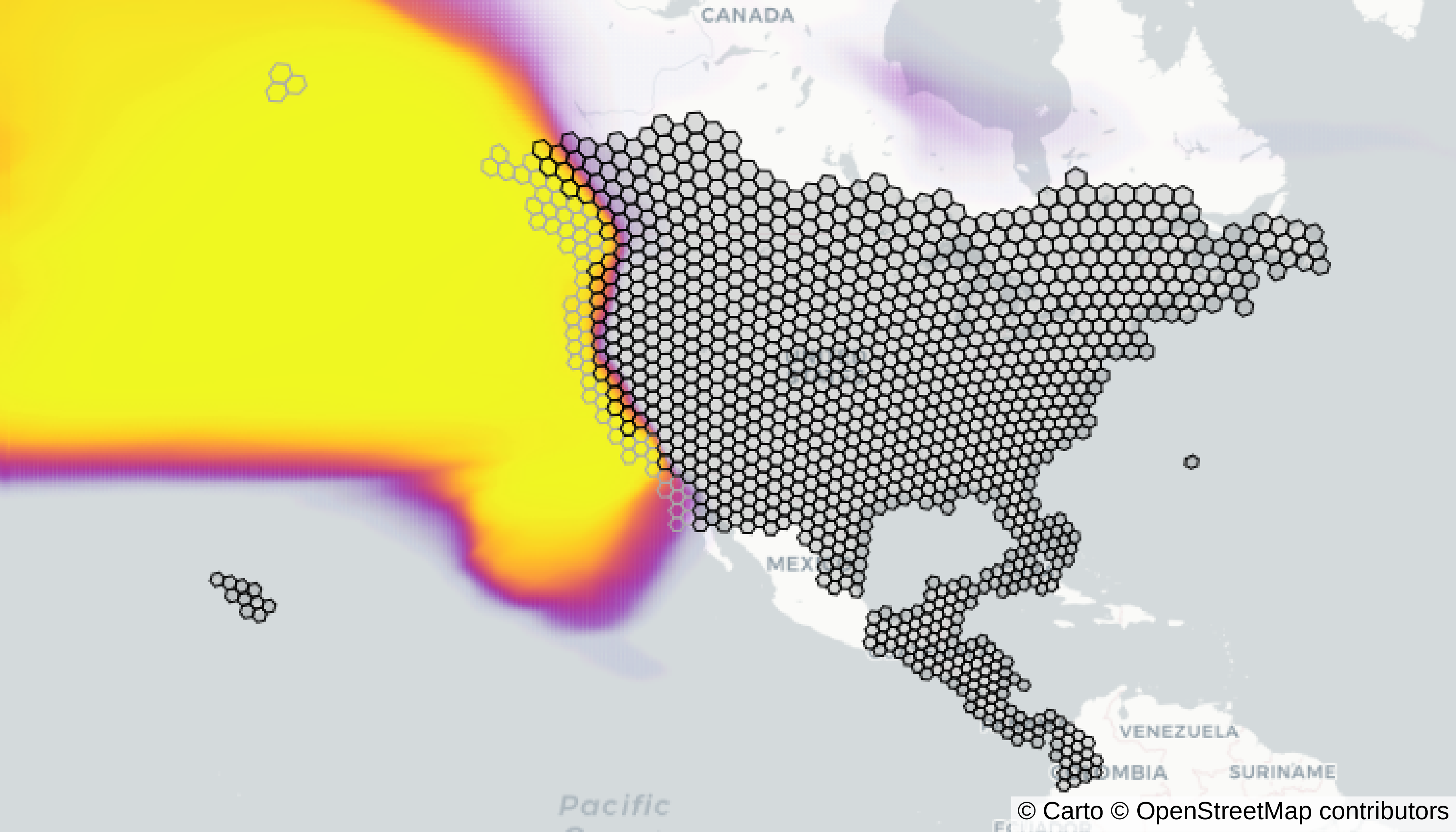} & \includegraphics[width=0.28\textwidth]{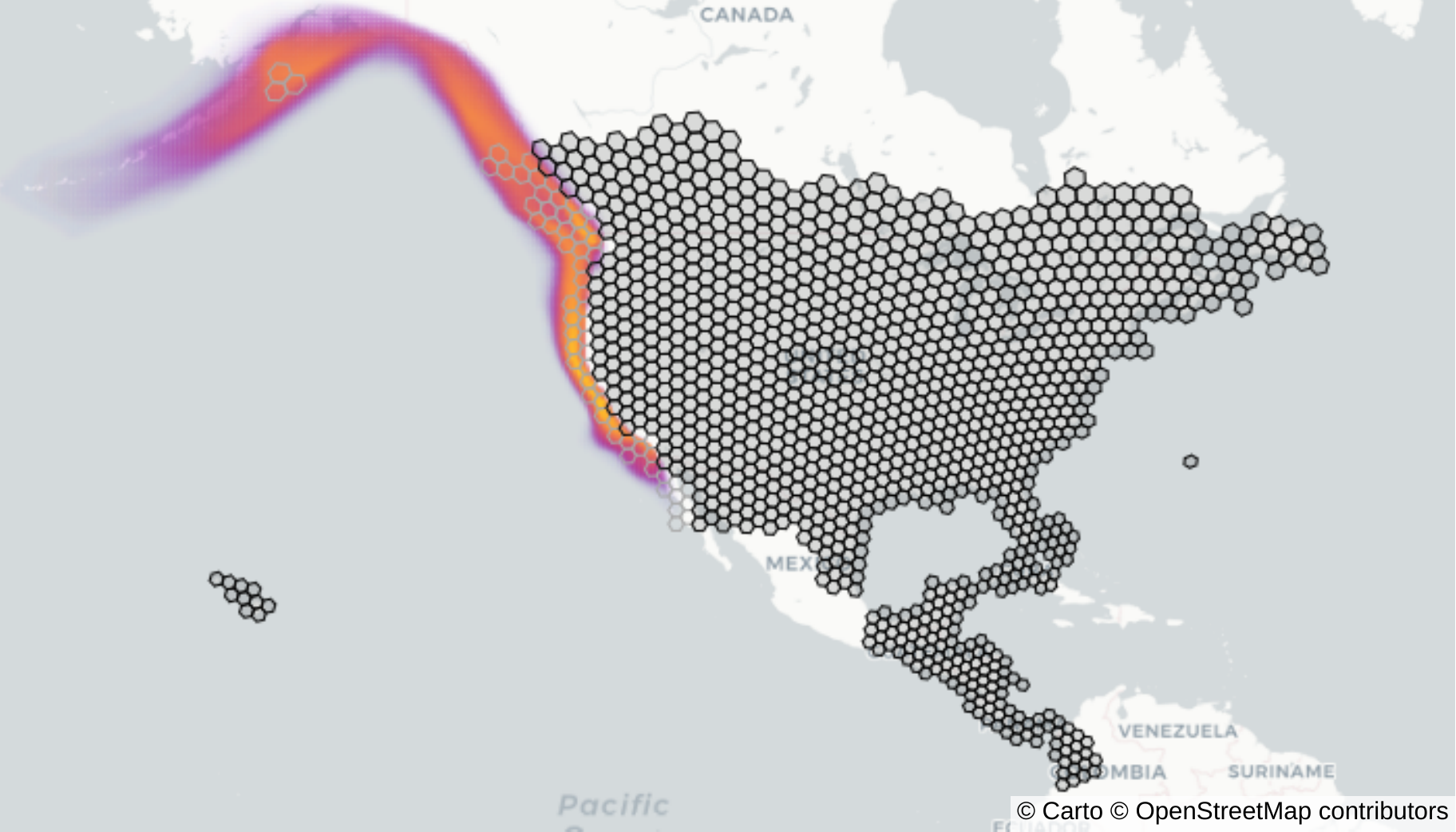} \\
\end{tabular}
}
\includegraphics[trim={0pt 0pt 0pt 0pt},clip,height=20px]{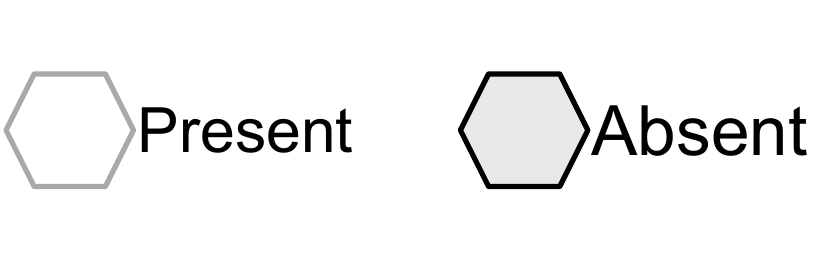}
\hspace{5pt}
\includegraphics[trim={0pt 0pt 0pt 0pt},clip,height=15px]{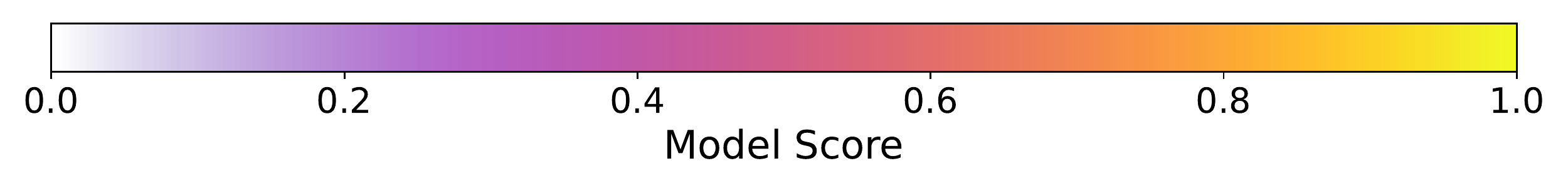} 
\vspace{-5pt}
\caption{
    Loss function comparison (columns) for three different species of birds (rows). All models were trained with 1000 examples per class. See Figure~\ref{fig:woothr_comparison} for an explanation of the plots. These species were chosen for visualization because their ranges have interesting complementary properties. (\emph{Top Row}) \href{https://ebird.org/species/barswa}{Barn Swallow} is a species that occurs across the globe. (\emph{Middle Row}) \href{https://ebird.org/species/burthr}{Blue Rock-Thrush} is a species whose range goes from the data rich \emph{Western Palearctic biogeographic realm}, through a data sparse area, and back to a more data rich area of eastern Asia. (\emph{Bottom Row}) \href{https://ebird.org/species/blkoys}{Black Oystercatcher} is a species whose range hugs the west coast of the United States. Unlike in other visualizations, these maps for Black Oystercatcher do not use a mask to filter out predictions from non-land regions. Here, we specifically wanted to see if the models learned to follow the coastline. We observe that $\mathcal{L}_\mathrm{AN-SLDS}$ incorrectly expands the range into the Pacific. All models use coordinates as input. 
}
\label{fig:three_bird_comparison}
\end{figure}

To build some intuition for the behavior of $\mathcal{L}_\mathrm{AN-SSDL}$, $\mathcal{L}_\mathrm{AN-SLDS}$, and $\mathcal{L}_\mathrm{AN-full}$, we compare these losses on three species that are known to have interesting ranges in Figure~\ref{fig:three_bird_comparison}. 

\section{Additional Discussion}
\label{sec:supp_discussion}

\subsection{How do the benchmark tasks proposed in this paper compare to existing SDM benchmarks?}\label{sec:supp_discuss_benchmarks}

Presence-only SDM is notoriously tricky to evaluate~\citep{beery2021species}, and there are few public benchmark datasets available for the task. Here we will discuss the two most relevant lines of prior work that have approached this evaluation problem (one from the ecology community and one from the machine learning community), and discuss where our benchmark is similar and different. 

To the best of our knowledge, \citet{elith2006novel} was the first attempt to systematically compare presence-only SDM algorithms across many species and locations. That work compared 16 SDM algorithms on a collection of taxonomically-specific datasets from 6 different regions, covering a total of 226 species. Presence-only data was used for training and presence-absence data was used for evaluation. Unfortunately the data was not made publicly available until~\citet{elith2020presence}. There are two main issues with this benchmark. First, the benchmark is not suitable for studying large-scale joint SDM. It has a small number of species overall, and there are at most 54 covered in any region. Second, the species in the dataset are anonymized. This makes it impossible to use their dataset to study large-scale SDM, because we cannot increase the size of their training with external data nor can we evaluate our trained models on their test data. 

Another line of work comes from the GeoLifeCLEF series of datasets and competitions~\citep{botella2018overview, botella2019overview, deneu2020overview, cole2020geolifeclef, lorieul2021overview, lorieul2022overview}. These benchmarks represent an attempt to scale up presence-only SDM, with the 2022 dataset covering 17k species with 1.6M species observations the U.S. and France. As in our benchmark, all of their training data is drawn from community science projects. The primary limitation of the GeoLifeCLEF benchmarks is that they use spatially biased presence-only data at test time, evaluating the problem as an information retrieval task instead of a spatial prediction task. 

Our benchmark can be viewed as a significant expansion of the GeoLifeCLEF line of work. Instead of being geographically limited to France and the U.S., we allow data from anywhere in the world. (See Figure~\ref{fig:snt_iucn_dist} for a visualization of the spatial coverage of the \emph{S\&T} and \emph{IUCN} tasks.) Instead of evaluating with presence-only data, we use presence-absence data like~\citet{elith2020presence}. However, unlike~\citet{elith2020presence}, we work at a large scale that allows us to study data scaling in SDM. Our indirect evaluation tasks (\emph{Geo Prior} and \emph{Geo Feature}) add complementary dimensions to presence-absence evaluation, and have no counterpart in~\citet{cole2020geolifeclef} or~\citet{elith2020presence}. 

\subsection{Environmental Covariates vs. Coordinates}\label{app:env_vs_coords}

One important characteristic of any SDM is whether or not it is \emph{spatially explicit}. Spatially explicit SDMs include geospatial coordinates as part of the model input~\citep{domisch2019spatially}. Traditional covariate-based SDMs include only environmental features (\eg altitude, distance to roads, average temperature, etc.) in the input~\citep{elith2009species}. 

Covariate-based SDMs are often understood to reflect \emph{habitat suitability}, because they learn a relationship between environmental characteristics and observed species occurrence patterns. A covariate-based SDM will make the same predictions for all locations with same covariates, even if those locations are on different continents. Furthermore, covariates sets must be selected by hand and are often limited in their spatial resolution and coverage. 

By contrast, spatially explicit SDMs can model the fact that a species may be present in one location and absent in another, even if those two locations have similar characteristics. However, spatially explicit models are unlikely to generalize to locations that are spatially distant from the training data -- such locations are simply out of distribution. 

SINRs trained with coordinates are spatially explicit, so our goal is not to learn from data in one location and extrapolate to distant locations. Instead, our goal is to use abundant (but noisy and biased) species observation data to approximate high-quality expert range maps. Our locations of interest are the same during training and testing. The difference is the training data source and quality. See \citet{merow2014we} for a more nuanced discussion of extrapolation vs. interpolation and the role of model complexity in SDM.

\subsection{The Role of Time}

Some species are immobile (\eg trees), while others (\eg birds) may occupy different areas at different times of the year. For this reason, there has long been interest in the temporal dynamics of species distributions~\citep{collins1991importance, guisan2011sesam}. However, traditional SDMs use environmental features as input, which seldom include temporal structure~\citep{elith2020presence, norberg2019comprehensive}. For instance, the popular WorldClim bioclimatic variables used in many SDM papers are non-temporal~\citep{hijmans2005very}. It is therefore not unusual for papers on SDM to make no explicit considerations for temporal information. Similarly, in this work we do not use temporal information during training or evaluation. However, we consider this to be an interesting area for future work. 

\section{Implementation Details}
\label{sec:impl_det}

\subsection{Network Architecture}

\begin{figure}[h]
\centering
\includegraphics[width=0.6\textwidth]{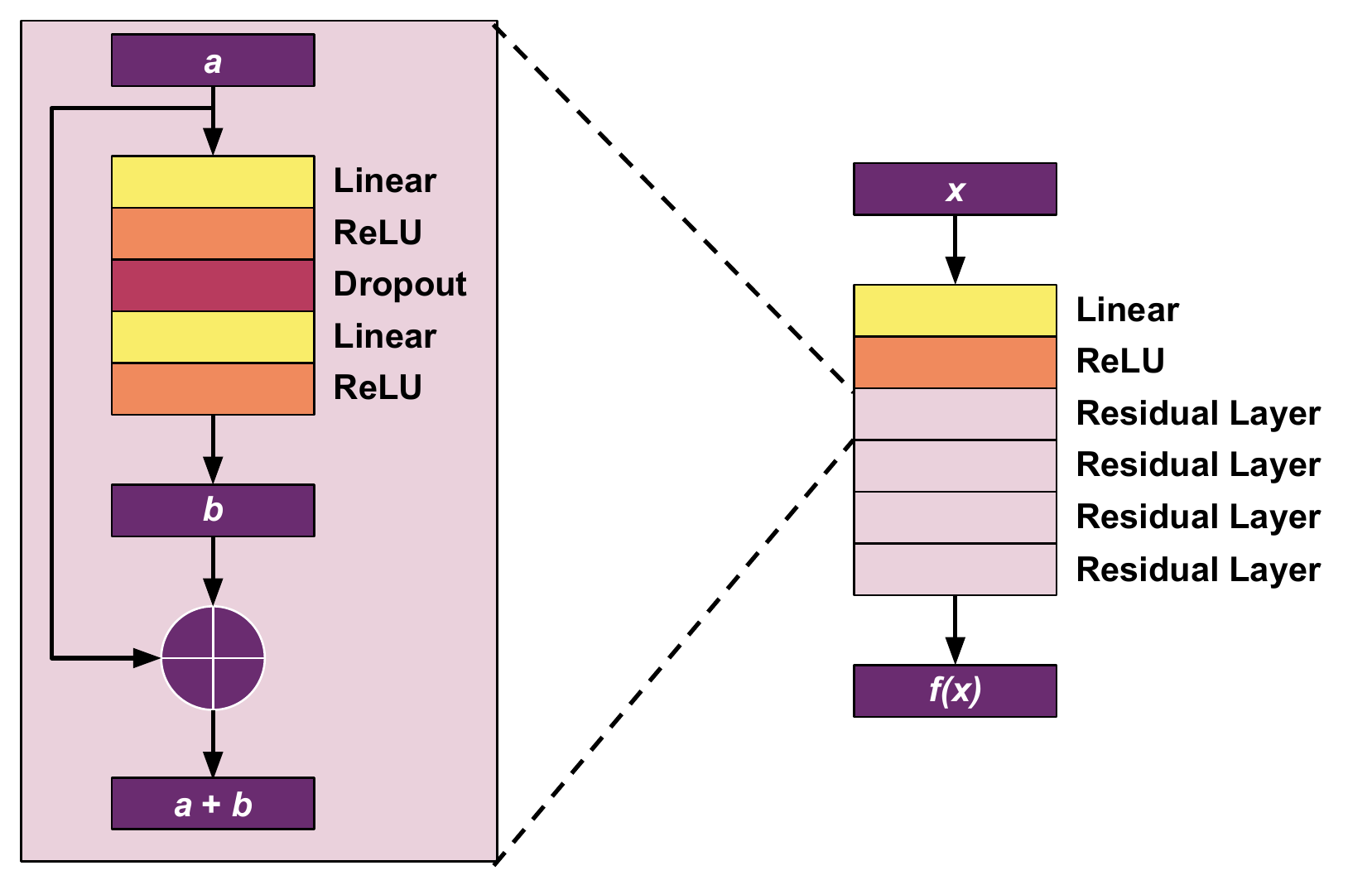}
\caption{
    Network diagram for the fully connected network (with residual connections) which we use for our location encoder $f_\theta$.
}
\label{fig:arch}
\end{figure}

We use the network in Figure~\ref{fig:arch} for our location encoder $f_\theta$. This is identical to the architecture in~\citet{mac2019presence}, and similar architectures have been used for other tasks~\citep{martinez2017simple}. The right side of the figure shows the network structure, consisting of one standard linear layer and four residual layers. The left side of the figure shows the structure of a single residual layer. Note that all layers have the same number of nodes. Every layer of the network has 256 nodes and we use $p = 0.5$ for the dropout probability.

\subsection{Training Details}
\label{sec:training_details}

\textbf{Environment.}
All models were trained on an Amazon AWS \texttt{p3.2xlarge} instance with a Tesla V100 GPU and 60 GB RAM. The model training code was written in PyTorch (v1.7.0). 

\textbf{Hyperparameters.} 
All models were trained for 10 epochs using a batch size of 2048 and a learning rate of $5e-4$. We used the Adam optimizer with an exponential learning rate decay schedule of
\begin{center}
    \texttt{learning\_rate} = \texttt{initial\_learning\_rate}$\times$ \texttt{epoch}$^{0.98}$
\end{center} 
where $\texttt{epoch} \in \{0, 1, \ldots, 9\}$. 
For $\mathcal{L}_\mathrm{AN-full}$ and $\mathcal{L}_\mathrm{GP}$ we set $\lambda = 2048$.

\subsection{Environmental Features}\label{sec:env_covs}

When environmental features are required for model inputs, we use the elevation and bioclimatic rasters from WorldClim 2.1~\citep{fick2017worldclim} at the 5 arc-minute spatial resolution. We normalize each covariate independently by subtracting the mean and dividing by standard deviation (ignoring \texttt{NaN} values). We then replace \texttt{NaN} values with zeros \ie the new mean value. 

\subsection{Baselines}\label{sec:baselines}

\subsubsection{Logistic Regression}\label{sec:impl_det_log_reg}

This section discusses our implementation of logistic regression with environmental covariates.  
The architecture for this approach is equivalent to a SINR but replacing the location encoder $f_\theta$ with the identity function. Then we can in principle use any of our loss functions for training. All other training details follow Appendix~\ref{sec:training_details}. 

\subsubsection{Discretized Grid}
In this section we describe our discretized grid baseline for the \emph{S\&T}, \emph{IUCN}, and \emph{Geo Prior} tasks, which is a simple spatial binning method. Once we choose a resolution level, the H3 geospatial indexing library~\citep{H3Web} defines a collection of $W$ cells $\{H_1, \ldots, H_W\}$ that partition the globe. For instance, $W=$ 2,016,842 at resolution level five. We show discretized grid results for a few different resolution choices in Table~\ref{tab:baseline_results}. Below we describe the discretized grid baseline in more detail. 

For the \emph{S\&T} and \emph{IUCN} tasks, we can compute a score for any hexagon and species as follows:
\begin{enumerate}
    \item We compute the number of occurrences of species $j$ in hex $w$ as
        \begin{align}
            n_{wj} &= \sum_{i=1}^{N} \indic{\mathbf{x}_i \in H_w} \indic{z_{ij} = 1}
        \end{align}
    for $1 \leq w \leq W$ and $1 \leq j \leq S$. 
    \item Let $H_{t}$ be a hexagon we wish to evaluate at test time. For any species $1 \leq j \leq S$, we compute a prediction for $H_t$ as
        \begin{align}
            \hat{y}_{j} = \frac{n_{tj}}{\max_{1 \leq w \leq W} n_{wj} }.
        \end{align}
\end{enumerate}
That is, $\hat{y}_j$ measures how often species $j$ was observed in $H_t$ (relative to how often species $j$ occurred in the location where it was observed most often). These predictions always fall between 0 and 1, which ensures that they are compatible with the average precision metrics we use for \emph{S\&T} and \emph{IUCN} evaluation. 

For the \emph{Geo Prior} task, the first step is the same but the second step is different: 
\begin{enumerate}
    \item We compute the number of occurrences of species $j$ in hex $w$ as
        \begin{align}
            n_{wj} &= \sum_{i=1}^{N} \indic{\mathbf{x}_i \in H_w} \indic{z_{ij} = 1}
        \end{align}
    for $1 \leq w \leq W$ and $1 \leq j \leq S$. 
    \item Let $H_{t}$ be a hexagon we wish to evaluate at test time. For any species $1 \leq j \leq S$, we compute a prediction for $H_t$ as
        \begin{align}
            \hat{y}_{j} = \indic{n_{wj} > 0}.
        \end{align}
\end{enumerate}
That is, any species which were not observed in $H_t$ are ``ruled out" for the downstream image classification problem. 

\begin{table*}[t]
\centering
\caption{
    Discretized Grid baseline results on test data when using various hexagon resolution for ``training" the model. As this baseline does not learn a location encoder, it is not possible to evaluate on the \emph{Geo Feature} task. 
\vspace{3pt}
}
\label{tab:supp_baseline_test}
\begin{tabular}{|l|c| c | c | c |}
\hline
    \multicolumn{2}{|c}{} & \multicolumn{1}{|c|}{Species Range} & \multicolumn{1}{|c|}{IUCN} & \multicolumn{1}{c|}{Geo Prior} \\ \hline
     Hex Res & \# / Cls. & MAP & MAP & $\Delta$ Top-1 \\ \hline
     $0$ & All & 54.67 & 21.89 & 3.5  \\
     $1$ & All & 61.56 & 37.13 & 4.1  \\
     $2$ & All & 61.03 & 36.92 & 3.1  \\
     $3$ & All & 51.09 & 26.57 & -0.9  \\
     \hline 
\end{tabular}
\label{tab:baseline_results}
\end{table*}

\section{Training Dataset}
\label{sec:sup_train_data}

\subsection{Dataset Construction}

Our training data was collected by the users of the community science platform iNaturalist~\citep{iNatWeb}. iNaturalist users take photographs of plants and animals, which they then upload to the platform. Other users review these images and attempt to identify the species. The final species labels are decided by the consensus of the community. Each species observation consists of an image and associated metadata indicating when, where, and by whom the observation was made. iNaturalist data only contains presence observations, \ie we do not have access to any confirmed absences in our training data.  

\begin{figure}[t]
\centering
\includegraphics[width=0.9\textwidth]{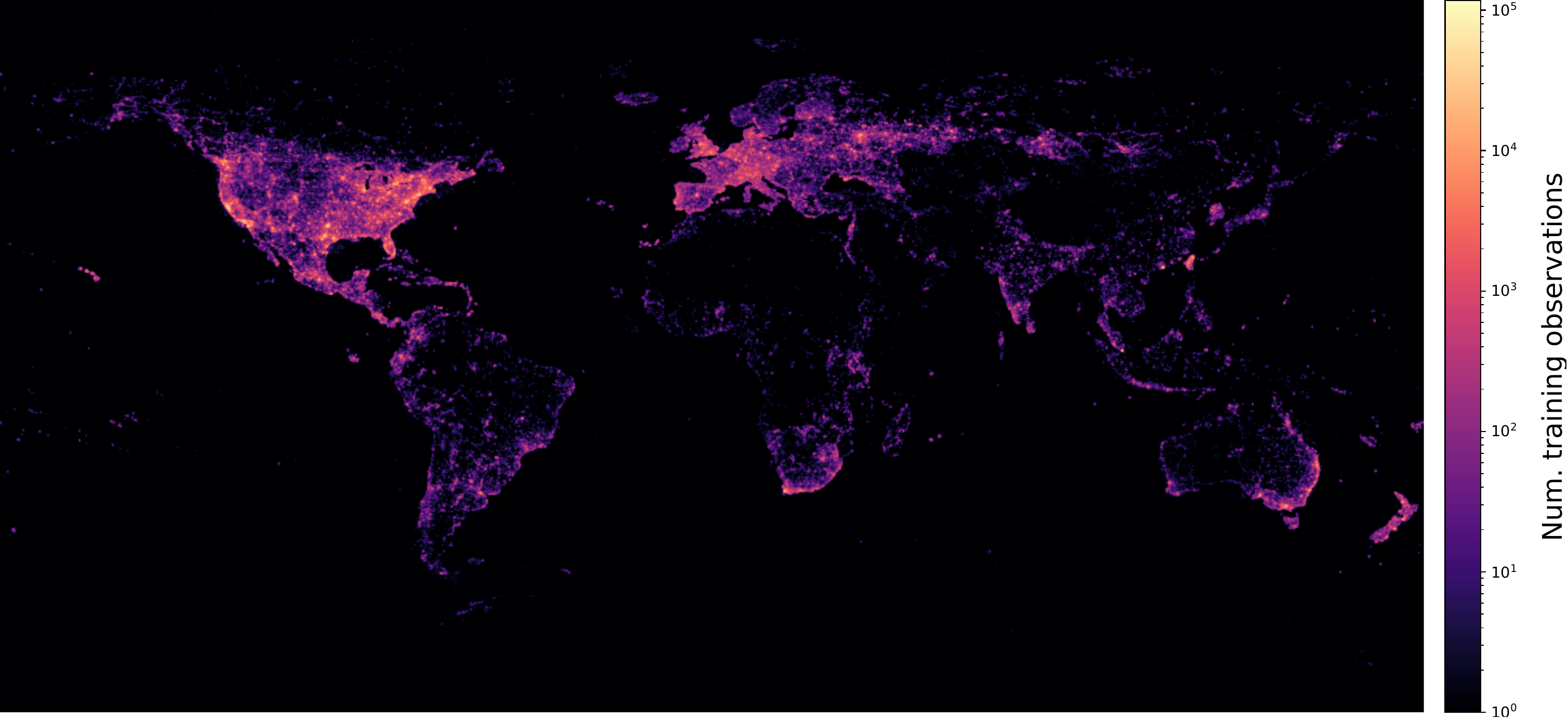}
\vspace{-5pt}
\caption{
    Histogram of the locations of the observations from our iNaturalist training set. Here we bin the data for all 35 million observations across all 47,375 species. Darker colors indicate fewer observations, brighter colors indicate more. The training data is biased towards North America, Europe, and New Zealand. 
}
\label{fig:inat_train}
\end{figure}

\begin{figure}[t]
\centering
\includegraphics[width=0.45\textwidth]{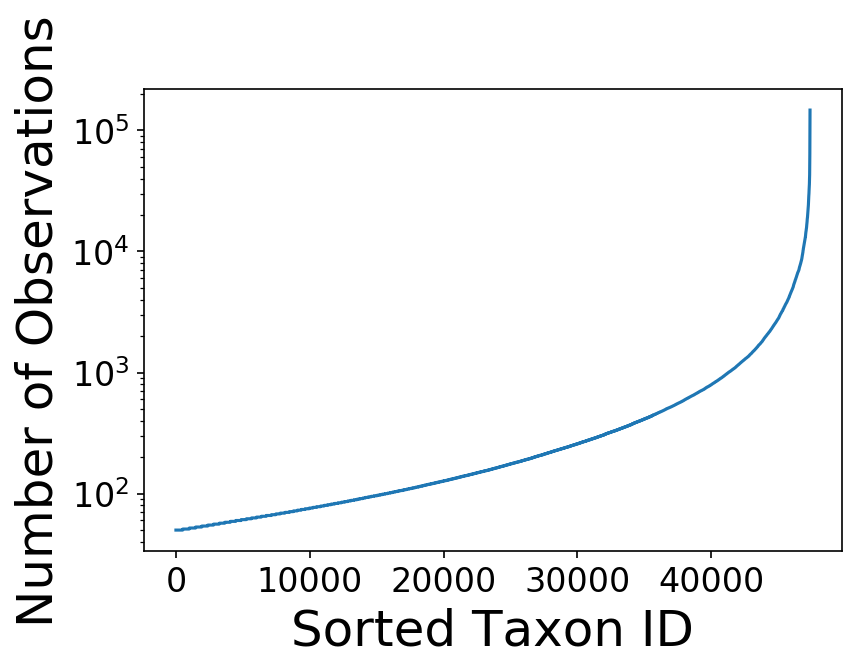}
\includegraphics[width=0.45\textwidth]{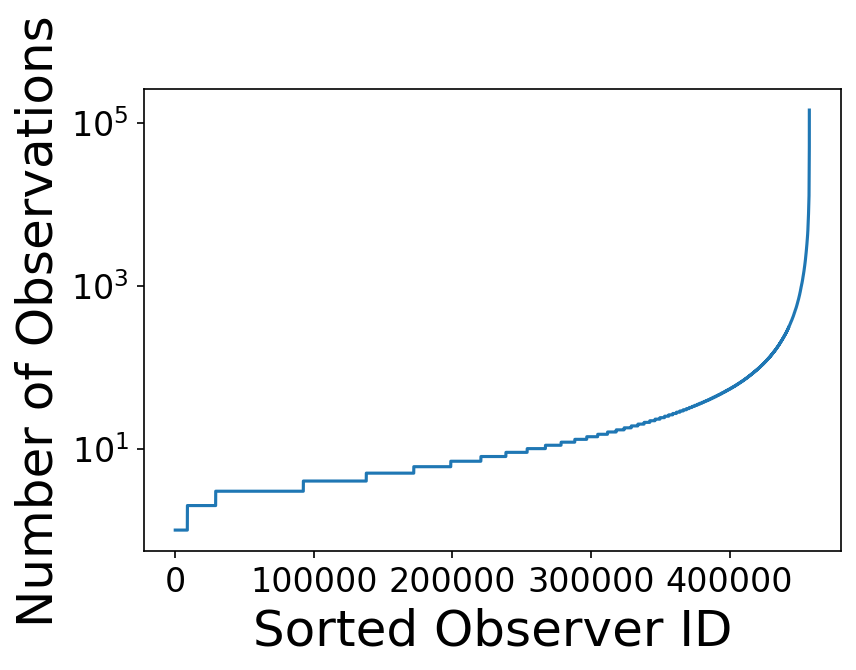}
\vspace{-5pt}
\caption{
    Summary statistics for our training observations data from iNaturalist. (\emph{Left}) Distribution of observations over species. (\emph{Right}) Distribution of observations over users (\ie observers). 
}
\label{fig:obs_stats}
\end{figure}

Specifically, our training data was sourced from the iNaturalist AWS Open Dataset\footnote{\href{https://github.com/inaturalist/inaturalist-open-data}{https://github.com/inaturalist/inaturalist-open-data}} in May 2022. 
This public split of the data does not include location data for sensitive species if they are deemed to be threatened by location disclosure. 
We began by filtering the species observations according to the following rules:
\vspace{-10pt}
\begin{enumerate}
    \itemsep-3pt 
    \item Observations must have valid valid longitude and latitude data.
    \item Observations must be identified to the \emph{species} level by the iNaturalist community. Observations which can only be identified to coarser levels of specificity are discarded. 
    \item Observations must have \emph{research grade} status, which indicates that there is a consensus from the iNaturalist community regarding their taxonomic identity.
\end{enumerate}
\vspace{-10pt}
After this filtering process, species with fewer than 50 observations were removed from the dataset. 
We also remove any species which are marked as \emph{inactive}\footnote{\href{https://www.inaturalist.org/pages/how+taxon+changes+work}{https://www.inaturalist.org/pages/how+taxon+changes+work}}.
Finally, we only included observations made prior to 2022. This will enable a temporal split from 2022 onward to be used as a validation set in the future. 
After filtering, we were left with 35,500,262 valid observations from 47,375 distinct species. 
A visualization of the geographical distribution of the resulting data can be seen in Figure~\ref{fig:inat_train}. As Figure~\ref{fig:obs_stats} shows, our training data is heavily imbalanced, reflecting the natural frequency with which they are reported to iNaturalist~\citep{van2018inaturalist}.

\subsection{Changing the Number of Training Examples per Category}

In the main paper we consider the impact of the number of \emph{observations} per species in the training set by training on different sub-sampled datasets. 
We construct these datasets by choosing $k$ observations per species, uniformly at random. 
We set a seed to ensure that we are always using the same $k$ observations per category. We also make certain that sampled datasets are nested, so the dataset with $k_1$ examples per category is a superset of the dataset with $k_2 < k_1$ examples per category. If a category has fewer than $k$ observations, we use them all.  

\subsection{Changing the Number of Training Categories}

In the main paper we also consider the impact of the number of \emph{species} in the training set. 
In particular, we consider the following nested subsets:
\vspace{-10pt}
\begin{itemize}
    \itemsep-3pt 
    \item The set of 535 bird species in the eBird Status \& Trends dataset \citep{fink2020ebird}. 
    \item The eBird Status \& Trends species plus an additional $A$ randomly selected species, where $A \in \{4000, 8000, 16000,  24000\}$.
\end{itemize}

\section{Evaluation Tasks}
\label{sec:sup_eval_data}

\begin{figure}[tb]
\centering
\includegraphics[width=0.8\textwidth]{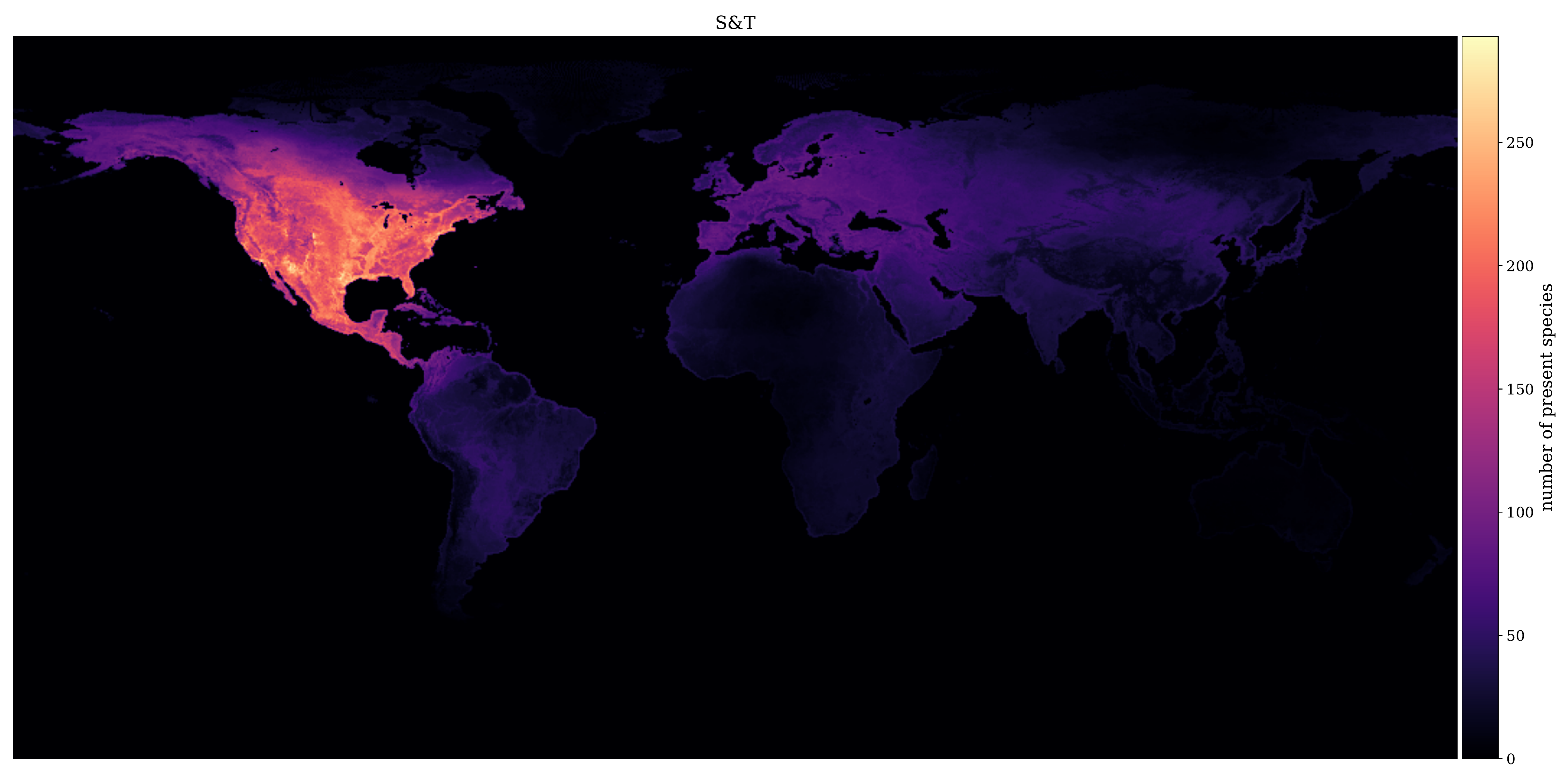}\\
\includegraphics[width=0.8\textwidth]{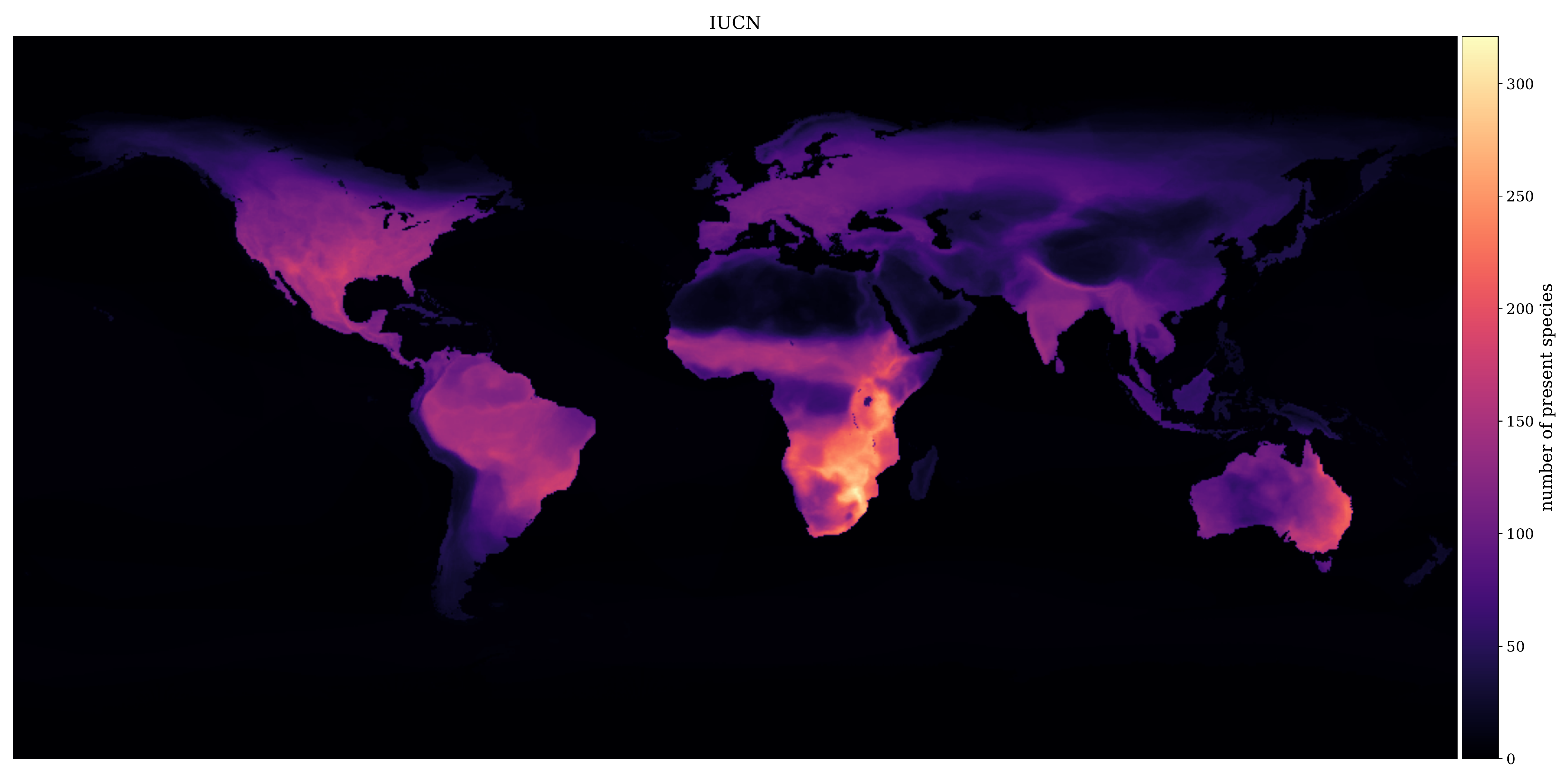}
\vspace{-5pt}
\caption{
    Visualization of the number of species present at different locations for the \emph{S\&T} task (top) and the \emph{IUCN} task (bottom).  Darker colors indicate fewer species, brighter colors indicate more species. The \emph{IUCN} task has much broader coverage than the \emph{S\&T} task. 
}
\label{fig:snt_iucn_dist}
\end{figure}

Here we provide additional details on the benchmark tasks used in the main paper. 
For each task, we outline the dataset properties, how it was collected, and the evaluation metrics used. In Figure~\ref{fig:snt_iucn_dist} we visualize the spatial coverage of the \emph{S\&T} and \emph{IUCN} tasks. 

\subsection{S\&T: eBird Stats and Trends Range Maps} 
\noindent\textbf{Task:} 
The goal of this task is to evaluate the effectiveness of models trained on noisy crowdsourced data from iNaturalist for predicting species range maps. 
We use the \textit{eBird Status \& Trends} data from~\citet{fink2020ebird} to evaluate our range predictions. 
This dataset consists of estimated relative abundance maps for 535 species of birds predominately found in North America, but also other regions. 
The relative abundance maps are computed at a spatial resolution of $3\times3$ km. 
The predictions are the output of an expert crafted model~\citep{fink2020ebird} that has been trained on tens of millions of presence and absence observations, makes use of additional expert knowledge to perform data filtering, and uses rich environmental covariates as input. 
While not without its own limitations, we treat this data as the ground truth for evaluation purposes as it is developed using much higher quality data and expert knowledge compared to what we use to train our models. 

\noindent\textbf{Dataset:}
We first download the rasterized abundance data for each species for all weeks of 2021 using the eBird API. 
We next reprojected each species' raster stack into latitude and longitude coordinates. 
We then spatially binned the data using H3 hexagons (\ie cells) at resolution five\footnote{\href{https://h3geo.org}{https://h3geo.org}}. 
2,016,842 cells cover the world at this resolution, each with an average area of $252.9\text{km}^2$. 
We finally sum all the relative abundance values for each cell, for each week of the year, for each species. 
Cells with nonzero values are considered present locations, cells with zero values are considered absent locations. 

Our goal is to predict the presence or absence of a given species in each hexagon using the \textit{eBird Status \& Trends} output as the (psuedo) ground truth.
The evaluation regions are restricted to those where the \textit{eBird Status \& Trends} models have determined that there is sufficient data to make a prediction for a given species. 
Thus, the set of evaluation regions can vary from species to species. 
For example,  \texttt{Melozone aberti} has 127,270 locations with known presence or absence, of which 549 are deemed present. 
On the other hand, \texttt{Columba livia} has 499,406 locations with known presence or absence, of which 132,807 are deemed present. 

The \textit{eBird Status \& Trends} data provides species presence and absence information for each location over the course of the year. 
For the purposes of our evaluation, we collapse the time dimension and count a hexagon region as being a presence for a given bird if the output of their model is greater than zero for any week in the year for that species.  

\noindent\textbf{Evaluation:}
We use mean average precision (MAP) for evaluation, only evaluating on valid regions for a given species.  

\subsection{IUCN: Range Maps}\label{sec:iucn_data}
\noindent\textbf{Task:}
The goal of this task is similar to the previous one, \ie to predict the geographical range of a set of species. 
However, instead of target range maps that are estimated by another model, here we use expert curated range maps (encoded as geospatial polygons) from the International Union for Conservation of Nature (IUCN)~\citep{IUCNRedListData}. 
This set of data contains a more taxonomically and geographically diverse set of species compared to the \textit{Stats and Trends} task, as it contains mammals, reptiles, and amphibians, in addition to birds. 
The bird data in this task comes from BirdLife International~\citep{BirdLifeData}. 
The IUCN data is from the `2022-2 update", last updated on the 9th of December 2022, and the BirdLife data is the ``2022.2" version. 

\noindent\textbf{Dataset:}
Of the 47k species in our training set, we first exclude all species where more than 10\% of the iNaturalist observations fall outside of the expert defined ranges and where there a taxonomic difference between IUCN and iNaturalist. 
This leaves 2,418 species that overlap with our training set.  
The data is contains 1,368 birds, 438 reptiles, 330 mammals, and 282 amphibians. 
Note our filtering cannot account for false positive regions from the IUCN data as we have no mechanism of extracting true absence from the iNaturalist source data. 

\begin{figure*}[t]
\centering
\includegraphics[width=0.95\textwidth]{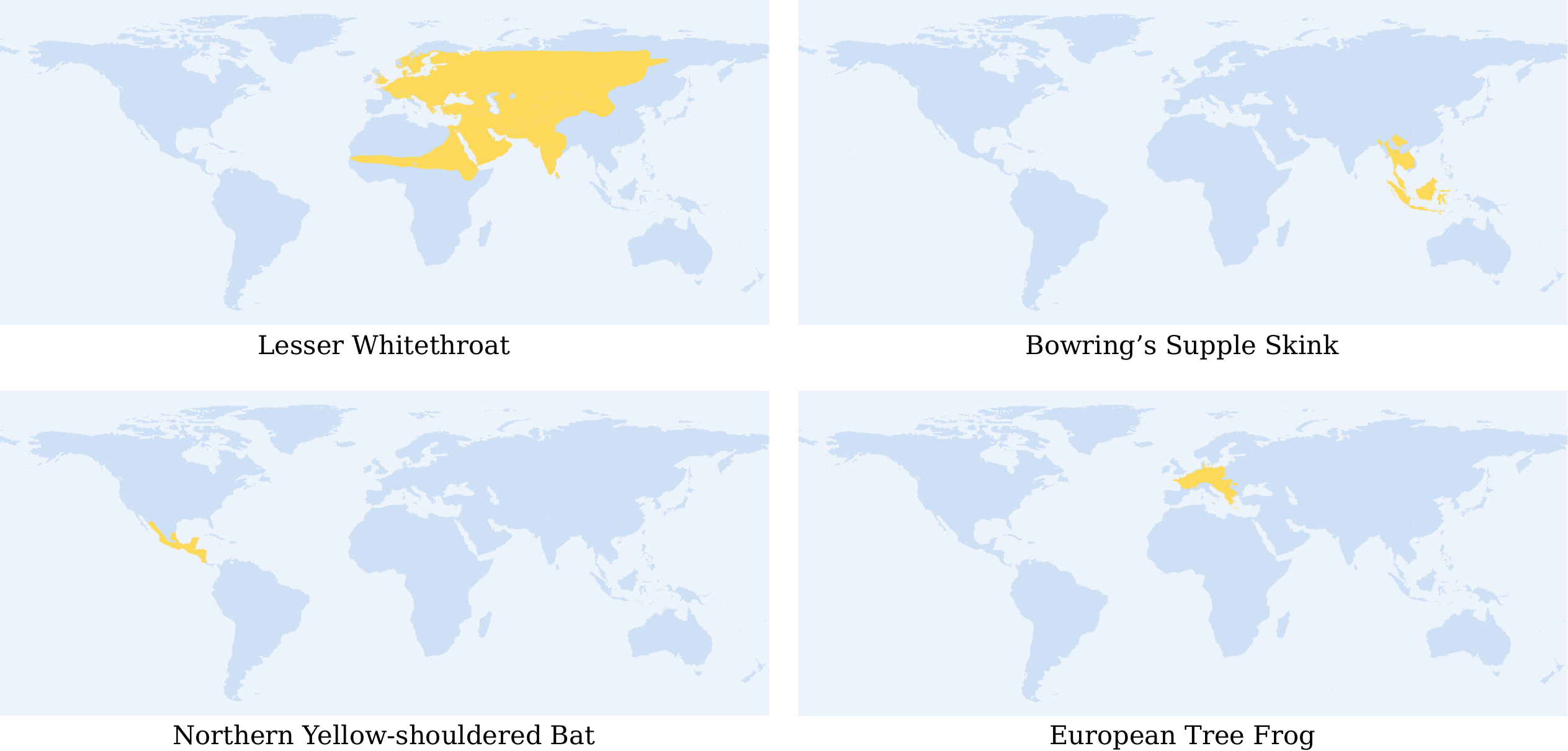}
\vspace{-1pt}
\caption{
    Expert defined ranges for four randomly selected species from our IUCN range evaluation task. The yellow regions indicate locations where the species is said to be present, otherwise they are absent. Light blue and darker blue indicate ocean and land respectively, and are only included for visualization purposes.
}
\label{fig:iunc_exs}
\end{figure*}

Using the H3 geospatial indexing library~\citep{H3Web}, we sample all locations (\ie latitude and longitude coordinates)  at resolution five to determine if each location is contained within the IUCN range polygon(s) for a given species. 
This results in 2,016,842 locations for each species, where each location denotes the centroid of the corresponding H3 cell. 
Each location is either marked as a true presence (if the cell centroid is contained within an IUCN polygon(s)) or a true absence (if the cell is \emph{not} contained within a polygon).
Note, these expert range maps cannot necessarily be assumed to be the objective ``ground truth'' (\ie species ranges can shift over time), but serve as strong proxy for it. 
A visualization of the expert provided ranges for a subset of species is shown in Figure~\ref{fig:iunc_exs}.  

\noindent\textbf{Evaluation:}
As for the \emph{S\&T} task, we use mean average precision (MAP) as the evaluation metric, which results in a single score for a model averaged across all species. 

\subsection{Geo Prior: Geographical Priors for Image Classification} 
\noindent\textbf{Task:} 
The goal of this task is to combine the outputs of the models trained for species range estimation on the iNaturalist dataset with computer vision image classifier predictions. 
This evaluation protocol has also been explored in other work, \eg \citet{berg2014birdsnap,mac2019presence}. 
We simply weight the probabilistic image classifier predictions for a given image with the species presence predictions from the location where that image was taken. 
The intuition is that the range prediction reduces the probability of a given species being predicted by the vision model if the range estimation model predicts that the species is \emph{not} likely to be present at that location. 

\noindent\textbf{Dataset:} 
For the vision classifier, we use an image classification model developed by the iNaturalist community science platform~\footnote{\href{https://www.inaturalist.org/blog/63931-the-latest-computer-vision-model-updates}{https://www.inaturalist.org/blog/63931-the-latest-computer-vision-model-updates}}. 
This model is an Xception network~\citep{chollet2017xception} that has been trained on 55,000 different taxonomic entities (\ie classes) from over 27 million images. 
We take the predictions from the final classification layer of the classifier, and do not apply any of their sophisticated taxonomic post-processing. 
There are a total of 49,333 species in the set of 55,000 classes -- the others are higher levels in the taxonomy, \eg genera. 
The images used to train the image classifier come from observations that were added to iNaturalist prior to December 2021. 

We then constructed a test set consisting of all research grade observations (\ie those observations for which there is a consensus from the iNaturalist community as to which species is present in the image). 
The images in the test set only contain the set of species that were observed at training time, \ie we do not consider the open-set prediction problem.  
The observations were selected from between January and May 2022 to ensure that they did not overlap with the training set. 
We take at most ten observations per species, which results in 282,974 total observations from 39,444 species. 
In practice, many species do not have 10 observations. 
In total there are 2,721 species (with 9,808 total images) that are not present in our range estimation training set. 
For each of the 282,974 observations, we extract the predictions from the deep image classifier across all 39,444 remaining species. 

\noindent\textbf{Evaluation:} 
Performance is evaluated in terms of top-1 accuracy, where the ground truth species label is provided by the iNaturalist community. 
Without using any information about where an image was taken, the computer vision model alone obtains an accuracy of 75.4\%, which increases to 90.4\% for top-5 accuracy. 
During evaluation, if a species is not present in a range model, we simply set the output for the range model for that species to $1.0$. 

\subsection{Geo Feature: Environmental Representation Learning} 
\noindent\textbf{Task:}  
This task aims to evaluate how well features extracted from deep models trained to perform species range estimation can generalize to \emph{other} dense spatial prediction tasks. 
Unlike the other benchmark tasks that use the species occupancy outputs directly, this is a transfer learning task. We remove the classification head $h_\phi$ and evaluate the trained location encoder $f_\theta$ in terms of downstream environmental prediction tasks.  
The intuition is that a model that is effective at range estimation may have learned a good representation of the local environment. 
If so, that representation should be transferable to other environmental tasks with minimal adaptation. 

This task is inspired by the linear evaluation protocol that is commonly used in self-supervised learning, \eg \citet{chen2020simple}. 
In that setting, the features of the backbone model are frozen and a linear classifier is trained on them to evaluate how effective they are on various downstream classification tasks. 
In our case, instead of classification, we aim to \emph{regress} various continuous environmental properties from the learned feature representations of our range estimation models. 
A related evaluation protocol was recently used in~\citet{rolf2021generalizable} for the case of evaluating models trained on remote sensing data. 

\begin{figure}[t]
\centering
\includegraphics[width=1.0\textwidth]{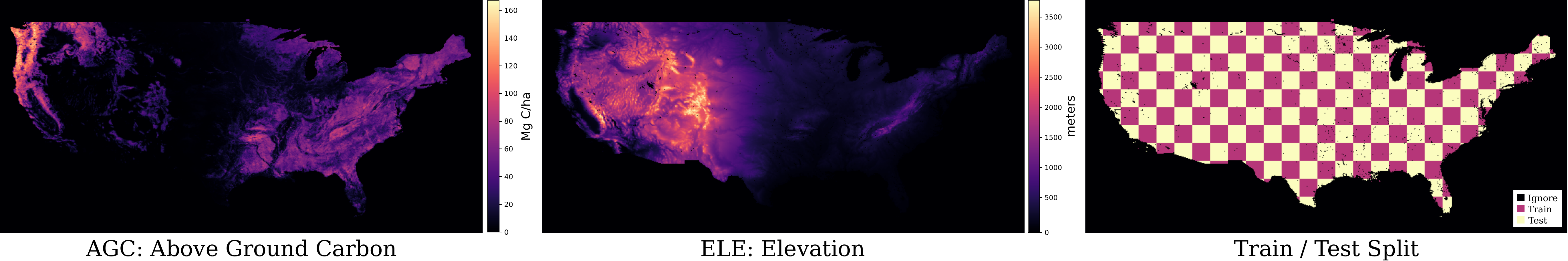}
\vspace{-15pt}
\caption{Here we illustrate two of the nine evaluation layers used in the \emph{Geo Feature} prediction task (left and middle). On the right we indicate which regions contain pixels that are in the train or test split, or simply ignored during evaluation. }
\label{fig:geofeats}
\end{figure}

\noindent\textbf{Dataset:} 
The task contains nine different environmental data layers which have been collected using Google Earth Engine~\footnote{\href{https://earthengine.google.com}{https://earthengine.google.com}}. 
The nine data layers are described in Table~\ref{tab:geofeats}. 
For each of the layers, we have rasterized the data so that the entire globe is represented as a $2004 \times 4008$ pixel image. 
Each pixel represents the measured value for a given layer for the geographical region encompassed by the pixel. 
Example images can be seen in Figure~\ref{fig:geofeats}. 

\noindent\textbf{Evaluation:} 
For evaluation, we crop the region of interest to the contiguous United States and grid it into training and test cells. 
The spatial resolution of the training and testing cells are illustrated in Figure~\ref{fig:geofeats} (right). 
Note, we simply ignore locations that are not in the training or test sets, \eg the ocean. 
This results in 51,422 training points and  50,140 test points. 
Features are then extracted from the location encoder for the spatial coordinates specified in the training split, and then a linear ridge regressor is trained on the train pixels and evaluated on the held out test pixels. 
The input features are normalized to the range $[0, 1]$.
We cross validate the regularization weighting term $\alpha$ of the regressor on the training set, exploring the set $\alpha \in
\{0.1, 1.0, 10.0\}$.
Performance is reported as the coefficient of determination $R^2$ on the test pixels, averaged across all nine layers. 

\begin{table}[h]
\centering
\caption{
    Description and sources of the nine environmental spatial layers that are part of our \emph{Geo Feature} prediction task.
}
\resizebox{1.0\linewidth}{!}
{
\begin{tabular}{|l | p{8cm} |l | l|} 
 \hline
 \textbf{Name} & \textbf{Task} & \textbf{Units} & \textbf{Range}\\  
 \hline 
AGC & Above ground living biomass carbon stock density of combined woody and herbaceous cover in 2010. \texttt{NASA/ORNL/biomass\_carbon\_density/v1} - \texttt{agb}& Mg C/ha &  0 to 129 \\  \hline
ELE & GMTED2010: Global multi-resolution terrain elevation data 2010. Masked to land only. \texttt{USGS/GMTED2010} - \texttt{be75} & meters  & -457 to 8746 \\ \hline
LAI & The sum of the one-sided green leaf area per unit ground area. \texttt{JAXA/GCOM-C/L3/LAND/LAI/V2} - \texttt{LAI\_AVE} - \texttt{2020}&   (leaf area per ground area) &  	0 to 	65531 \\ \hline
NTV & Percent of a pixel which is covered by non-tree vegetation. \texttt{JAXA/GCOM-C/L3/LAND/LAI/V2} - \texttt{LAI\_AVE} - \texttt{20202}  & \% & 0 to 100\\ \hline
NOV & Percent of a pixel which is not vegetated. \texttt{MODIS/006/MOD44B} - \texttt{Percent\_NonVegetated} & \% & 0 to 100 \\ \hline
POD & UN adjusted estimated population density. \texttt{CIESIN/GPWv411/GPW\_UNWPP-Adjusted} \texttt{\_Population\_Density} - \texttt{unwpp-adjusted\_population\_density} & \# of persons / km$^2$ & 0 to 778120 \\ \hline
SNC & Normalized difference snow index snow cover. \texttt{MODIS/006/MOD10A1} - \texttt{NDSI\_Snow\_Cover} - \texttt{2019} & (amount snow cover) & 0 to 100 \\ \hline
SOM & Soil moisture, derived using a one-dimensional soil water balance model. \texttt{IDAHO\_EPSCORTERRACLIMATE} - \texttt{soil} - \text{2020} & mm &  0 to	8882\\ \hline
TRC & The percentage of pixel area covered by trees. \texttt{NASA/MEASURES/GFCC/TC/v3} - \texttt{tree\_canopy\_cover} - \texttt{2000-2020} & \% & 0 to 100 \\ \hline
\end{tabular}
}
\label{tab:geofeats}
\end{table}

\section{Reproducibility Statement}

The information needed to implement and train the models outlined in this paper is provided in Appendix~\ref{sec:impl_det}. 
In addition, the different training losses we study are described in Section~\ref{sec:main_losses}. Training and evaluation code is available at: 
\begin{center}
    \url{https://github.com/elijahcole/sinr}
\end{center}

\section{Ethics}  

This work makes use of species observation data provided by the iNaturalist community. 
We only use the public data exports from iNaturalist, ensuring that sensitive data (\eg data related to species at risk of extinction) is not used by our models. 

As noted in the limitations section in the main paper, extreme care must be taken when attempting to interpret any species range predictions from the models presented in this paper. Our work is intended to provide (i) a proof-of-concept for large-scale joint species distribution modeling with SINRs and (ii) benchmarks for further model development and analysis. However, our models have failure modes and our benchmarks have blind spots. Further validation is necessary before using these models for conservation planning or other consequential use cases.